\documentclass[10pt,journal]{IEEEtran}
\usepackage{xcite}
\usepackage{xr-hyper}

\usepackage{algcompatible}

\ifCLASSINFOpdf
\usepackage{orcidlink}
   \usepackage{graphicx}
\else
\fi
\usepackage{xcolor}
\usepackage{svg}

\usepackage{amsmath}
\usepackage{slashed}
\usepackage{amssymb}
\usepackage{bm}
\usepackage{amsfonts}
\usepackage{subfiles}

\usepackage{algorithm2e}

\usepackage{url}
\usepackage{fancyhdr}

\RestyleAlgo{ruled}

\begin{document}
%
\title{Hyperspectral Unmixing Hierarchies}
%
%

\author{Joseph L. Garrett\,\orcidlink{0000-0001-8265-0661},~\IEEEmembership{Member,~IEEE,}
P. S. Vishnu, Pauliina Salmi, Daniela Lupu, Nitesh Kumar Singh,\\ Ion Necoara, Tor Arne Johansen\,\orcidlink{0000-0001-9440-5989},~\IEEEmembership{Senior Member,~IEEE,}
\thanks{J. L. Garrett (joseph.garrett@ntnu.no), P. S. Vishnu, T. A. Johansen from Department of Engineering Cybernetics, Norwegian University of Science and Technology }
\thanks{P. Salmi from Department of Biology, Norwegian University of Science and Technology}
\thanks{D. Lupu, and I. Necoara from Department of Automatic Control and Systems Engineering, University Politehnica Bucharest.}
\thanks{N. K. Nitesh from School of Computer Science
UPES, Dehradun, India.}}

%
%

\markboth{Hyperspectral Unmixing Hierarchies - To be submitted to IEEE for possible publication}%
{Garrett \MakeLowercase{\textit{et al.}}: }
%



\maketitle

\begin{abstract}
Unmixing reveals the spatial distribution and spectral details of different constituents, called endmembers, in a hyperspectral image.
Because unmixing has limited ground truth requirements, can accommodate mixed pixels, and is closely tied to light propagation, it is a uniquely powerful tool for analyzing hyperspectral images. 
However, spectral variability inhibits unmixing performance, the proper way to determine the number of endmembers is ambiguous, and the clarity of the endmembers degrades as more are included.
Hierarchical structure is a possible solution to all three problems.

Here, hierarchical unmixing is defined by imposing a hierarchical abundance sum constraint on Deep Nonnegative Matrix Factorization.
Binary Linear Unmixing Tactile Hierarchies (BLUTHs) solve the hierarchical unmixing problem with a simple network architecture. 
Sparsity modulation unmixing growth tailors the topology of a BLUTH to each scene.
The structure imposed by BLUTHs allows endmembers with varying levels of spectral contrast to be revealed, mitigating the challenge of spectral variability.

The performance of BLUTHs exceeds state-of-the-art unmixing algorithms on laboratory scenes, particularly with regard to abundance estimation, while their performance remains competitive on remote sensing scenes. 
In addition, ocean color unmixing by BLUTHs is demonstrated on hyperspectral scenes from the HYPSO and PACE satellites.
\end{abstract}


\IEEEpeerreviewmaketitle

\section{Introduction}

\begin{figure}
\centering

\includegraphics[width=0.48\textwidth]{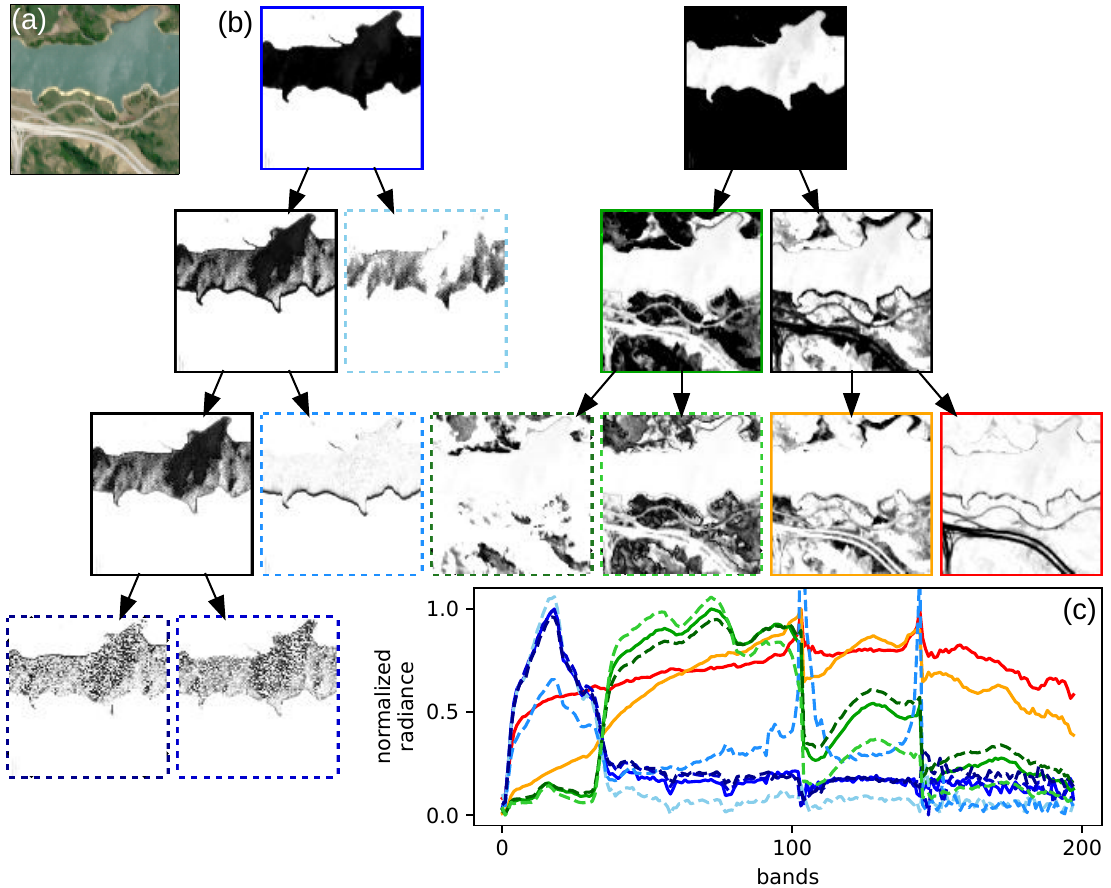}
\caption{Although the Jasper scene (a) is commonly partitioned into four endmembers in order to evaluate hyperspectral unmixing techniques (b, colored edges), the data contain a more detailed structure, which is depicted at four levels of an unmixing hierarchy. The finely partitioned spectra show subtle differences (c). Solid lines indicate endmembers present in the labels and dashed lines show finely-partitioned endmembers.}
\label{fig:Jasper_deep_structure}
\end{figure}

\IEEEPARstart{H}{yperspectral} imaging, occasionally called imaging spectroscopy \cite{goetz_imaging_1985}, collects detailed pixel-wise information at finely-spaced wavelengths about the response of a scene to incident electromagnetic radiation. It is used to study coatings\cite{raeissi_detection_2022,ma_coating_2023}, microplastics \cite{primpke_rapid_2020,faltynkova_use_2024}, geology \cite{dumke_first_2018,van_exem_new_2019,sahoo_modelling_2023}, artwork \cite{catelli_can_2018}, food \cite{badaro_near_2021}, and remote sensing from drones and satellites \cite{goetz_imaging_1985,zhang_uav_2025}, among other applications. 
The recently launched Plankton, Aerosol, Cloud, ocean Ecosystem (PACE) satellite images the whole Earth daily, laying the foundation for consistent long-term monitoring of large-scale ecosystem dynamics \cite{dierssen_synergies_2023}, while several small satellite hyperspectral constellations, such as HYPSO, Zhuhai-1, and Hyperfield \cite{grotte_ocean_2021,zhang_-orbit_2022,tikka_hyperfield_2023}, which regularly acquire higher-resolution imagery of a few locations, as well as demonstration missions such as EnMAP, PRISMA, and EMIT \cite{roger_high-resolution_2024,lou_variational_2025,reddy_integrating_2024}.  
Hyperspectral data are well-suited for the monitoring of aquatic environments, particularly the optically complex waters.
The analytic tools to process the images from the numerous and varied hyperspectral imaging satellites must be developed.

Unmixing identifies the constituents, or endmembers, present in a scene and estimates how the light reflected from a pixel is split between them \cite{bioucas-dias_hyperspectral_2012} (for example, in Fig. \ref{fig:Jasper_deep_structure}). 
The unmixing process consists of several tasks: the estimation of the number of endmembers, endmember extraction, and abundance estimation as well as according to the level of supervision they require \cite{cui_realistic_2023}. 
The effectiveness of unmixing has lead it to be used as the basis for many other procedures in hyperspectral image processing, such as multispectral-hyperspectral fusion \cite{yokoya_coupled_2012} and hyperspectral content-based image retrieval \cite{omruuzun_novel_2024}. 
Despite the successes of hyperspectral unmixing, there have been some recurrent problems, some of which plague every algorithm. 
First, algorithms must accommodate the spectral variation of endmembers, which is readily apparent even in the simplest of scenes \cite{somers_endmember_2011}. 
Second, unmixing algorithms generally require the number of endmembers to be determined in advance, but the various techniques for estimating the number of endmembers in advance generally give inconsistent estimates \cite{drumetz_spectral_2020}. 
These challenges are connected. 
For any given scene, if fewer endmembers are included in the scene, then there will be more spectral variation within endmembers. 

Ocean color (OC) is an ideal use-case for exploring how these challenges affect unmixing.
The main contributors to OC are phytoplankton, non-algal particles, including sediments, and dissolved organic matter (cDOM) \cite{spyrakos_optical_2018}.
Optically complex waters, such as inland waters, estuaries, and fjord systems, contain spectra that are dominated by sediments and cDOM with varying characteristics which causes water quality estimation algorithms to generalize poorly.
Moreover, the exact number of constituents is always unknown due to biological and topological variation (shifting shorelines, sediment characteristics, phytoplankton, dissolved material).
Because deep water tends to be very dark, small systematic effects caused by thin clouds or imperfect atmospheric compensation can have a notable effect. 
In addition, water presents rapidly changing spatial patterns and on-water measurements are valid only for a few hours. 
Unmixing can address the first challenge by revealing how land and cloud pixels contribute to the signal, called the adjacency effect \cite{burazerovic_detecting_2013}. It can mitigate the second by identifying endmember spectra so that the subsequent analysis can be connected to spectra themselves rather than specific geographic coordinates. 
Aquatic phenomena such as harmful algal bloom identification \cite{alfaro-mejia_blind_2023} and extent estimation \cite{pan_novel_2017}, mining pollution \cite{kopackova_applying_2014}, and algal species composition, both in nature \cite{legleiter_spectral_2022} and in a laboratory \cite{mehrubeoglu_resolving_2014,naik_blind_2025}, have all been explored with unmixing.
Moreover, unmixing is similar to optical water type (OWT) frameworks that cluster together observations of the water based on spectral measurements.
OWTs provide a basis for picking or blending subsequent algorithms based on optical conditions. 
OWTs have been created by techniques including by $k$-means clustering \cite{spyrakos_optical_2018}, fuzzy $c$-means clustering\cite{moore_optical_2014,atwood_framework_2024}, hierarchical clustering \cite{shi_classification_2014}, and gaussian processes \cite{blix_learning_2022}. 

Hierarchical structure has been used to adapt to spectral variation or to uncertainty in the number of endmembers.
Multiple Endmember Spectral Mixture Analysis (MESMA) introduced endmember bundles in a semi-supervised manner \cite{roberts_mapping_1998,bateson_endmember_2000}, while an automated variant allowed it to operate unsupervised \cite{somers_automated_2012}. 
Endmember bundles introduce 2 levels of hierarchy and increase the robustness of unmixing against spectral variation. 
MESMA has been applied in many applications such as algal bloom detection and wildfire burn area monitoring \cite{fernandez-manso_burn_2016,legleiter_spectral_2022,peterson_oil_2015,fernandez-garcia_multiple_2021}.
A similar group of methods replace the lowest level of the hierarchy with a probability distribution \cite{themelis_novel_2012,altmann_unsupervised_2014,chen_toward_2017,zhou_gaussian_2018}.

A few methods have extended the hierarchical structure beyond two levels. 
Hierarchical MESMA extends the endmember bundle concept up to 4 layers of depth with 20 endmembers, but still relies on partial supervision \cite{franke_hierarchical_2009}.
In Hierarchical Clustering of Hyperspectral Images Using Rank-Two Nonnegative Matrix Factorization (H2NMF), non-negative matrix factorization (NMF) clusters hyperspectral data into a binary decision tree and extracts representative endmember spectra \cite{gillis_hierarchical_2014}. 
Multilayer unmixing creates a hierarchy of abundances by applying NMF repeatedly, with the output of each layer as the input to the objective function of the subsequent layer \cite{rajabi_spectral_2015}. 
In contrast, Deep Nonnegative Matrix Factorization (DNMF) trains the layers simultaneously \cite{feng_hyperspectral_2018,fang_sparsity-constrained_2018,li_self-supervised_2022,huang_hyperspectral_2023}.
However, \cite{de_handschutter_consistent_2023} showed that the original update rule was inconsistent with the original DNMF objective function and proposed self-consistent data-centric and layer-centric versions of the objective functions.
A recent comparison suggested that although hierarchical methods were less accurate on a benchmark dataset, their robustness enhanced their applicability to real data from the PRISMA satellite \cite{settembre_advancing_2025}.

Here the Hierarchical Abundance Sum Constraint (HASC) is introduced and imposed on DNMF
so that the subsequent layers of the networks can be interpreted as continually finer distinctions between different endmember spectra. 

The primary contributions of this article are:
\begin{enumerate}
    \item A new hierarchical network, the Binary Linear Unmixing Tactile Hierarchy (BLUTH), demonstrates that the HASC can address spectral variability and endmember number uncertainty.
    \item The Sparsity Modulation Unmixing Growth (SMUG) algorithm grows BLUTH networks without a pre-defined structure.
\end{enumerate}
The primary contributions are built on two smaller developments which can be used independently:
\begin{enumerate}
\item[a)] Two variants of Archetypal Analysis \cite{cutler_archetypal_1994}  compatible with mini-batching.
\item[b)] Two variants of unmixing by annealing \cite{peng_nonnegative_2012}, one stochastic and one deterministic.
\end{enumerate}
The BLUTH networks are compared to other unmixing techniques on a series of 8 standard hyperspectral unmixing scenes and evaluated for ocean color unmixing on images of an algal bloom from the PACE and HYPSO satellites.

\section{Background}

\subsection{What is unmixing?}

Unmixing is the process of partitioning the pixels of a spectral image. 
The linear mixing model (LMM) describes unmixing under the assumption that photons only interact with a single material. Within the LMM, unmixing is the process of identifying the spectra of the endmembers (S) in an image and their abundances (A) in each pixel:
\begin{equation}
    Y \approx SA,
    \label{eq:lmmmodel}
\end{equation}
where $Y \in \mathbb{R}^{B\times N}$ is the original hyperspectral image, $S \in \mathbb{R}^{B\times P}$, $A \in \mathbb{R}^{P\times N}$, $P$ is the number of endmembers, $B$ the number of bands and $N$ the number of pixels. 
Two assumptions, phrased as constraints, permit unmixing to be interpreted in terms of light scattering. 
The first assumption is that the whole signal from a pixel is the sum of the signal from the individual endmembers in that pixel, called the Abundance Sum Constraint (ASC). 
The second assumption is that no endmember can cover a negative surface area, which is called the abundance non-negativity constraint (ANC).
Although more sophisticated models of light scattering can be accommodated, the linear mixing model with these two constraints remains central to unmixing and its interpretation.

\subsection{Varieties of unmixing}

Hundreds of variants have been developed because unmixing is both a necessary task in hyperspectral image processing but also challenging and nonconvex \cite{bioucas-dias_hyperspectral_2012}. A recent review \cite{feng_hyperspectral_2022} categorizes many of the algorithms, while the HySUPP experiments provide a clear numerical comparison of several methods, focusing on the degree of annotation \cite{rasti_image_2024}.

Unmixing algorithms can be categorized as Geometric, Non-negative Matrix Factorization (NMF) based, or deep learning based \cite{feng_hyperspectral_2022}.
This article focuses on NMF-based methods because they provide sufficient performance and are amenable to customization \cite{lee_learning_1999}. 
Unmixing algorithms also vary in the amount of supervision which they require: unsupervised (occasionally called blind), semi-supervised (e.g. with spectral libraries), or supervised (with labeled pixels) \cite{rasti_image_2024}.
Only unsupervised algorithms are tested here because they are suitable for remote scenes in which the specific endmembers are not known in advance. 
Moreover, while library-based unmixing would have benefits, it suffers worse performance in practice \cite{rasti_image_2024}.

\begin{figure}[t]
  \centering
  \includesvg[width=0.4\textwidth]{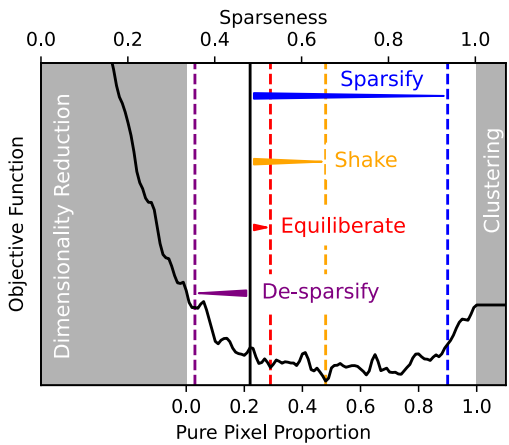}
  \caption{Unmixing techniques share characteristics with both dimensionality reduction (DR) and clustering, and those former techniques can be viewed as unmixing with the pure pixel proportion taken to the 0 (DR) or 1 (clustering) limit. Sparsity modulation unmixing growth (SMUG) grows a network by forcing it back and forth across this phase diagram. SMUG alternates between four modalities. The equilibrate modality (red) uses gradient descent to move the network from its current set of parameters (black) towards a local minimum of the objective function without adjusting the sparsity penalty. The sparsify modality (blue) gradually increases the sparsity penalty in order to push the network towards a pure pixel proportion setpoint. The de-sparsify modality (purple) forces the network to become less sparse.  The shake modality (orange) periodically applies a sparsity penalty to push the network out of insufficiently sparse local minima. Note that the objective function resides in a much higher dimensionality than the simplified 1D representation depicted here.}
  \label{fig:sparsity_phase_diagram}
\end{figure}

Because overly smooth output have been observed as a common failure mode of unmixing algorithms, sparsity-promoting regularization has become common, with over 30 variants listed in \cite{feng_hyperspectral_2022}.
Even though sparsity, or the number of elements of $A$ which vanish, is not differentiable, related penalties on the abundances or endmember spectra can be used. 
Sparseness, which refers to the ratio of the L1 and L2 norms scaled between 0 and 1, is a way to quantify sparsity within NMF \cite{hoyer_non-negative_2004,jia_constrained_2009}. 
Moreover, group sparsity has been applied the aforementioned endmember bundles, showing the compatibility of sparsity-promoting penalties with hierarchy \cite{drumetz_hyperspectral_2019}.
In the HySUPP experiments, it is reported that methods which apply sparsity regularization directly to abundances show worse performance \cite{rasti_image_2024}. 
The paradox of sparsity is that unmixing often produces scenes which are insufficiently sparse relative to manual labels, but directly incorporating sparsity into the objective function decreases performance. 

Other algorithms include sparsity regularization not as a penalty term in the objective function, but rather as a constraint. 
Many early geometrical methods involve a process for identifying pixels which contain non-zero abundance for only one endmember, called pure pixels \cite{chang_fast_2006}. 
More recently, Archetypal Analysis, in which all endmember spectra are convex sums of individual pixels, has become a simple way to induce sparsity \cite{cutler_archetypal_1994,akhtar_rcmf_2017,zouaoui_entropic_2023}. 

The objective functions used for unmixing are non-convex, which implies the presence of local minima. 
Entropic Descent Archetype Analysis (EDAA) addresses nonconvexity by running with an ensemble of initial conditions and selecting the optimal run based on secondary criteria \cite{zouaoui_entropic_2023}.
Markov random fields have been adopted as a tool for modeling the spatial dependencies. Simulated annealing and Hamiltonian Monte Carlo have been proposed as tools which allow them to escape local minima \cite{altmann_unsupervised_2014,chen_toward_2017}.
Similarly Deterministic Annealing NMF (DA-NMF) includes a term to decrease, rather than increase, the sparsity of the objective function, which, taken to its limit, makes the objective function convex \cite{peng_nonnegative_2012}. The non-convexity of the DA-NMF objective function is gradually introduced by decreasing the magnitude of the regularization.
While DA-NMF has a single minimum when regularized, the maximally sparse objective function would have $p^{N}$ degenerate local minima, illustrating the connection between sparsity and non-convexity.

A few algorithms aim to unmix larger datasets, although most unmixing algorithms are tailored for small test scenes of fewer than $10^{5}$ pixels.
The Distributed Parallel Geometric Distance method focuses on how to estimate the endmembers separately for subsection of the data so that they can later be fused \cite{canada_distributed_2025}. 
While hyperspectral scenes can be analyzed individually, they are often parts of much larger data repositories collected over many years.
Joint unmixing of multi-temporal scenes was shown to mitigate inconsistencies that were present if the scenes were unmixed separately \cite{henrot_dynamical_2016,thouvenin_online_2016}.
For example, FM-MESMA demonstrates how MESMA can be accelerated on multi-temporal image sequences \cite{borsoi_fast_2021}.
The experiments on larger dataset typically have focused on increasing the number of pixels analyzed, rather than the number of endmembers.
To increase the number of endmembers productively requires the imposition of some structure. 
Hierarchical unmixing is needed to equip the analysis of larger datasets with the capability of considering the relationship between constituents.

\subsection{Unmixing and other hyperspectral processing techniques}

Unmixing is related to other hyperspectral processing techniques \cite{hong_interpretable_2021}. 
Unmixing, in the limit that pure pixel proportion (PPP) of the scene vanishes, can be interpreted as dimensionality reduction,
while it can be interpreted as clustering in the opposite limit (Fig \ref{fig:sparsity_phase_diagram}).
The connection between clustering and unmixing has been exploited by a few techniques. 
Because H2NMF initializes unmixing with divisive hierarchical clustering, it implicitly adheres to the HASC.
The method of Veganzones et al, similarly initializes hierarchical clustering with unmixing, but, unlike H2NMF, does not attempt to extract endmembers from the clusters \cite{veganzones_hyperspectral_2014}. 
In a similar vein, agglomerative hierarchical clustering has been used to estimate the appropriate number of endmembers \cite{prades_estimation_2020}. 
In that work, initial clusters were formed with $k$-means clustering, because hyperspectral images are typically too large for fully agglomerative hierarchical clustering, as the computational time it requires scales as $O(N^{2})$ \cite{murtagh_algorithms_2012}. 

\section{Mechanics of Unmixing Hierarchies}

\subsection{Notation}

\begin{figure}
\centering
\includegraphics[width=0.45\textwidth]{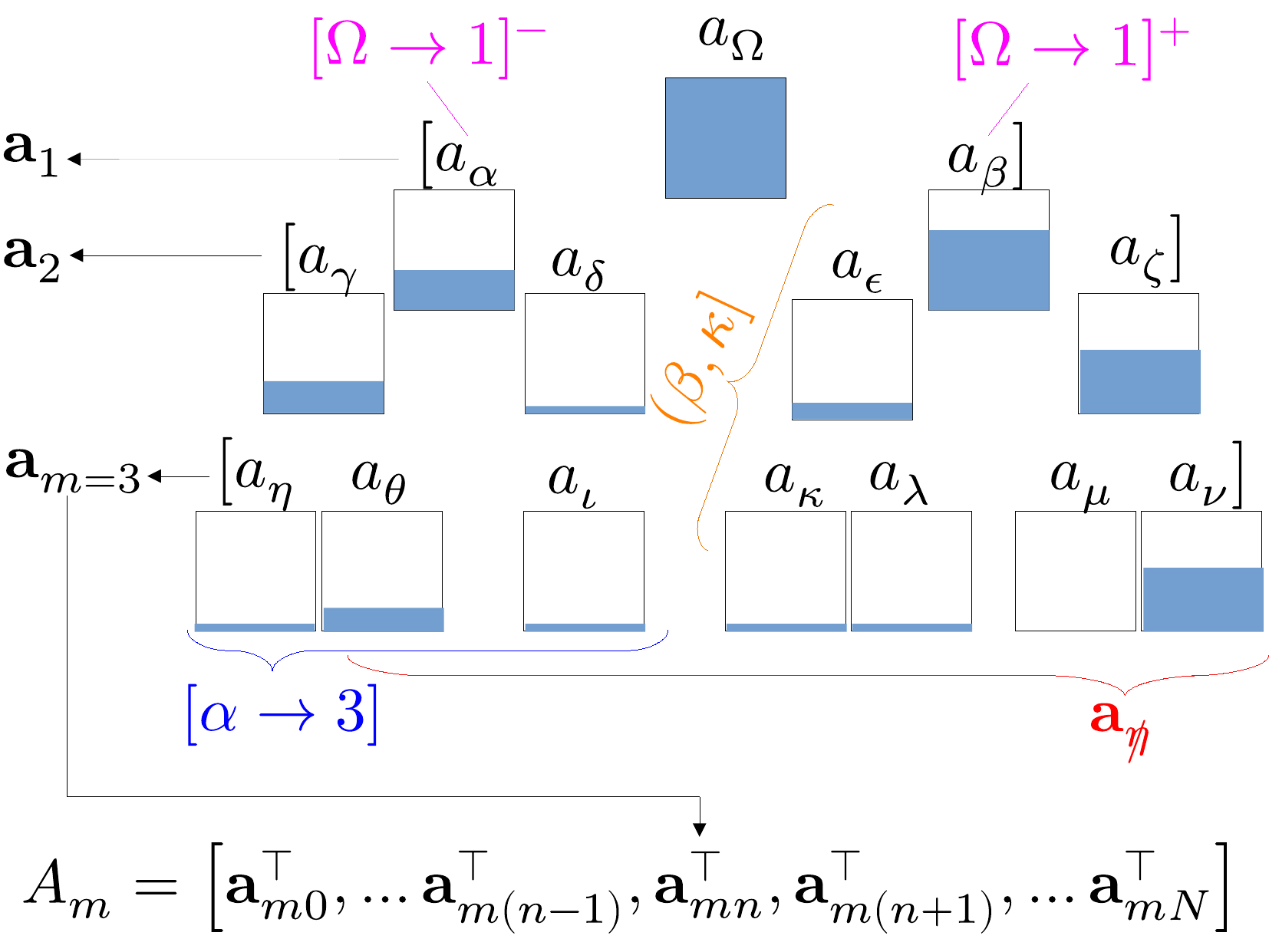}
\caption{Succinct notation is used for several groups of endmember abundances: the descendants of node $\alpha$ at level $m$ (blue), the endmembers on level $m$ complementary to $\eta$ (red), each side of a split (pink), the descendants of node $\beta$ which precede node $\kappa$ (orange), each are given their own notation. }
\label{fig:hUH_nomenclature}
\end{figure}

Matrices are represented as capital letters, vectors as bold lower-case letters, and scalars as lower-case letters. The endmember indices and regularization parameters will be indexed with Greek letters, pixels and bands will be denoted with the Latin alphabet. For simplicity of exposition, the individual endmember $\mathbf{s}_{\zeta}$ or spectrum $\mathbf{y}_{n}$ and abundance $\mathbf{a}_{n}$ of a each pixel will be used instead of the full matrices. The subscript $n$ is dropped whenever only a single pixels is considered. A few groups related to the hierarchy structure required additional notation (Fig. \ref{fig:hUH_nomenclature}). 

\subsection{What is an unmixing hierarchy?}

NMF searches for a factorization of a matrix $Y$ into two lower-rank matrices $S$ and $A$ by minimizing a objective function of the form $\|Y - SA\|_{\text{F}}^{2}$,
where $\|  \cdot \|_{\text{F}}$ is the Frobenius norm \cite{lee_learning_1999}. 
In contrast, hierarchical unmixing factors $Y$ into a series of matrices $S_{m}\in \mathbb{R}^{B\times k_{m}}$ and $A_{m}\in \mathbb{R}^{k_{m} \times N}$ which minimize $\sum_{m}^{M}\mu_{m}\|Y - S_{m}A_{m}\|_{\text{F}}^{2}$, akin to the data-centric objective proposed in \cite{de_handschutter_consistent_2023}, where $k_{m}$ is the number of endmembers at level $m$, and $\mu_{m}$ is the weight given to the $m$th level. 

The difference between hierarchical unmixing and the variants of DNMF proposed in \cite{de_handschutter_consistent_2023} comes from the coupling between subsequent $A_{m}$, rather than from the objective function itself. 
In DNMF, the feature matrices are coupled via matrix multiplication so that $A_{m-1}=U_{m} A_{m}$, where $U_{m}\in\mathbb{R}^{k_{m-1}\times k_{m}}$ is a non-negative matrix and the $k_{m}$ are sorted in ascending order. 
The $U_{m}$ are updated throughout the optimization process in DNMF, whereas the the $U_{m}$ are fixed in hierarchical unmixing except when the structure of the network changes. Then updates to $A_{M}$ implicitly update all of the abundances.  
For a single pixel, the HASC can be written as:
\begin{align}
a_{\alpha} =\sum_{\beta \in [\alpha \rightarrow m]} a_{\beta},
\label{eq:general_layer_coupling}
\end{align}
which, applied to all pixels, implies that $A_{M}$ determines the $A_{m}$ for $m<M$.
The HASC implies that the elements of $U_{m}$ must be 0 or 1, each column must sum to 1, and each row must sum to a positive integer (the number of nodes into which each endmember is split). 

The constraints which characterize the hierarchical unmixing objective function are phrased most simply as vectors.
The data-consistency term is $f(S,\mathbf{a})=\|\mathbf{y}-S\mathbf{a}\|_{2}^{2}$ for a single pixel at a single level.
A sparsity regularization term is added to the objective function as well, $g(\mathbf{a})=-\|\mathbf{a}\|^{2}_{2}$, which can be interpreted as the L1-L2 difference norm \cite{yin_ratio_2014}, with the L1 term dropped since it is constant.
The full optimization problem is then written as a sum over pixels $n$ and levels $m$:
\begin{align}
\min_{S_{m},{\bf a}_{mn}} \sum_{m}^{M}\sum_{n}^{N} \bigg[\mu_{m} f(S_{m},{\bf a}_{mn}) + \gamma_{m} g(\mathbf{a}_{mn})\bigg],
\label{eq:obj}
\end{align}
subject to (i) the abundance nonnegativity constraint (ANC), (ii) the abundance sum to 1 constraint (ASC) and (iii) the HASC. 
The ANC is phrased as $S_{m},\mathbf{a}_{mn}\geq0$, while $a_{\Omega n}=1$ suffices to express the ASC when the HASC is included, and $\Omega$ is the unitary endmember. 
Then $\gamma >0 $ promotes sparsity while $\gamma <0$ discourages sparsity.
Constraints of the form (iv) $S_{m}=YZ_{m}$ for $Z_{m}\in\mathbb{R}^{N\times k_{m}}$, with the constraints that (v) $Z_{m}\geq 0$ and (vi) $\sum_i Z_{mij}=1$, can be additionally included to enforce archetypal analysis. The final constraint, (vii) $Z_{mij}=0\text{ or }1$, is used to enforce pure pixel analysis (Appendix \ref{ap:spectra}).
In summary, basic hierarchical unmixing minimizes equation \ref{eq:obj} with constraints (i-iii), while the other constraints enable AA updates of the endmember spectra.

\subsection{Binary Unmixing}

Suppose there are only two endmembers: $\bm{s}_{\alpha}, \bm{s}_{\beta}\in \mathbb{R}^{B}$. 
Let $\bm{y}\in \mathbb{R}^{B}$ be a hyperspectral pixel. 
Then the endmember abundance $\mathbf{a}$ is determined by the $x \in [0,1]$ which minimizes
\begin{align}
    \|\bm{y} - (x \bm{s}_{\alpha} + (1-x) \bm{s}_{\beta})\|^{2}_{2} - \gamma \|\mathbf{a}\|_{2}^{2}.
    \label{eq:line}
\end{align}
where $\mathbf{a}=[x, \text{ } (1-x)]$.

If $\|\bm{s}_{\alpha}-\bm{s}_{\beta}\|^{2}_{2}>\gamma$, the minimizing $x$ can be written analytically:
\begin{align}
    x = \Phi \Bigg(\frac{\bm{w}^{\intercal} \bm{y} - d + 1}{2}\Bigg),
    \label{eq:lambda}
\end{align}
where $\bm{w}=2\bm{p}/(\|\bm{p}\|^{2}_{2}-\gamma)$, $\bm{p}=\bm{s}_{\alpha}-\bm{s}_{\beta}$, $d= \bm{w}^T (\frac{\bm{s}_{\beta}+\bm{s}_{\alpha}}{2})$, and $\Phi$ is a clipped rectified linear unit (cReLU) that truncates $x$ to the relevant range:
\begin{align}
    \Phi(u) = 
    \begin{cases}
    & 0 \text{, if } u \leq 0 \\
    & u \text{, if } 1 > u > 0 \\
    & 1 \text{, if } u \geq 1 \\
    \end{cases}.
\end{align}

\subsection{Binary Linear Unmixing Tactile Hierarchies}

\begin{figure}[t]
  \centering
  \includesvg[width=0.48\textwidth]{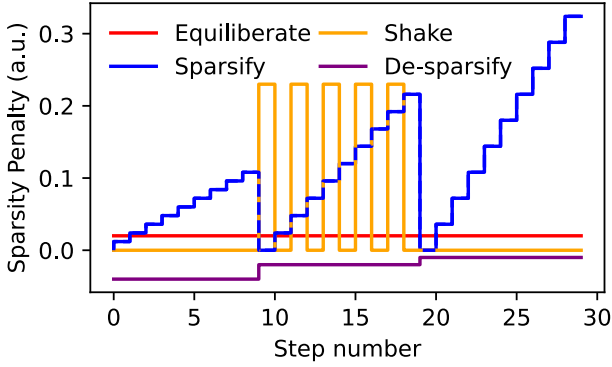}
  \caption{The equilibrate modality uses a constant sparsity penalty, $\gamma$ (red). The sparsify modality gradually increases the penalty within each set (3 sets of 10 steps pictured), but resets it between them (blue). In the shake modality, the original value of the objective function is evaluated in the first set, then a periodic sparsity penalty is applied in the second set, and the objective function around the new minima is evaluated in the third step (orange). During fine-tuning, the de-sparsify modality uses a negative penalty function (purple).}
  \label{fig:dynamic_penalties}
\end{figure}

The binary unmixing framework is extended to images with more than two endmembers through the use of a multilayer structure. 
A Binary Linear Unmixing Tactile Hierarchy (BLUTH) structures the endmembers in order to retain the clarity of 2-endmember unmixing. 
Rather than updating the abundances directly, BLUTHs update them implicitly through the weights defined in equation \ref{eq:lambda}.
Then the whole abundance map is implicitly updated each time the weights are updated, facilitating batching. 
The weights, $\mathbf{w}_{\zeta}$ and $d_{\zeta}$, rather than being calculated from the spectra of each pixel, are learned globally. 
The word {\it tactile} in BLUTH refers to how the weights of the network, although learned, can be interpreted through equation \ref{eq:lambda} as hyperplanes and modified directly. 

When the HASC is enforced on binary unmixing, the abundance of an endmember at level $m+1$ for a single pixel can be calculated from the abundance at level $m$:
\begin{equation}
a_{\zeta^{+}} = x_{\zeta^{+}}a_{\zeta} \text{ and }a_{\zeta^{-}} =(1-x_{\zeta^{+}})a_{\zeta}
\end{equation}
where $\zeta$ is an endmember with two direct endmember descendants, $\zeta^{+}$ and $\zeta^{-}$, and $x_{\zeta^{+}}$ is the splitting coefficient defined in equation \ref{eq:lambda}.
Extending this inductively leads to the BLUTH formula for each $a_{\zeta}$:
\begin{equation}
    a_{\zeta} = \prod_{\beta \in [\Omega,\zeta)}x_{\beta}(\bm{y}),
    \label{eq:abundance_product}
\end{equation}
where $\Omega$ is the origin endmember which contains all others as descendants. Since $a_{\zeta}(\bf{y})$ is a function of $\bf{y}$ while the weights are held constant, it is not strictly necessary to retain the entirity of $A$ in memory. 
Note that the normalization of $\bf{y}$ described in Appendix \ref{ap:normalization} applies only to the objective function equation \ref{eq:obj}.

\subsection{Training overview}

\begin{figure}
\centering
\includegraphics[width=0.48\textwidth]{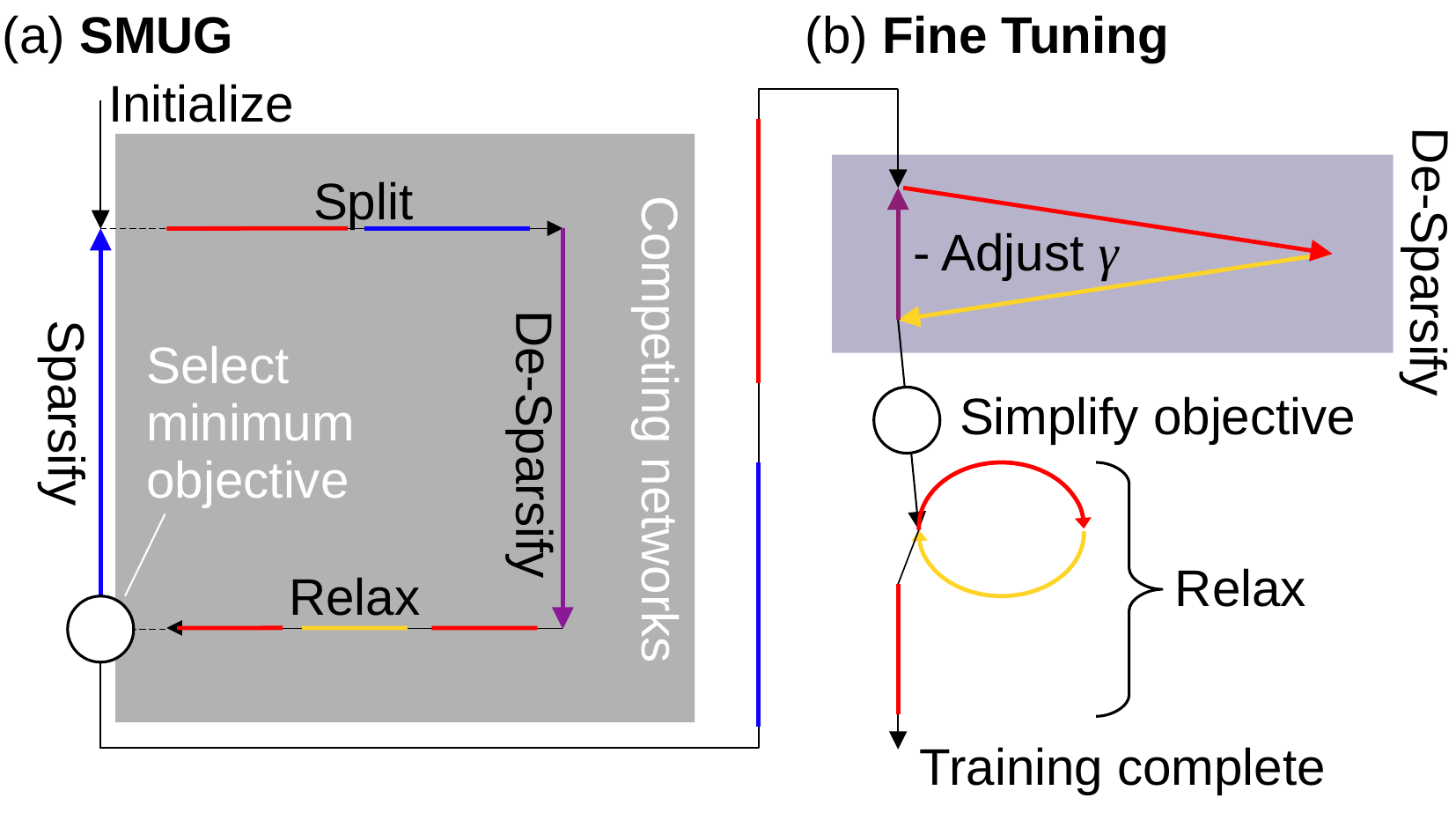}
\caption{The Sparsity Modulation Unmixing Growth procedure sequentially adds endmembers to the hierarchy, minimizing the objective function at each step (a). In the fine-tuning stage, the structure is held constant, while the endmembers relax (b). The colors indicate the modalities of Figure \ref{fig:dynamic_penalties}.}
\label{fig:hUH_process}
\end{figure}

The training of a BLUTH is partitioned into two stages: a growth stage and a fine-tuning stage (Fig. \ref{fig:hUH_process}). 
Both stages use an alternating minimization framework to separately update the abundances $A$ and the endmember spectra $S$.
The splitting weights $\mathbf{w}_{\zeta}$ are updated sequentially through block coordinate (gradient) descent strategy \cite{lange_optimization_2013}, and $A$ is updated implicitly through them.
At each step, the direction of the gradient is calculated first, and then a line search method finds the minimum of the objective function along that direction. 
The mathematical details of the abundance updates can be found in appendices \ref{ap:updates} and \ref{ap:numintegration}.

In the growth stage, the endmember spectra are selected from the spectra in the data, in a process here designated as Pure Pixel Analysis (PPA). PPA constrains the $S$ update step so that a spectrum which already represents an endmember cannot be selected again. For the tests reported here, the fine tuning stage is run twice. The first time uses PPA for spectra updates, as in the growth stage, while the second uses AA. Both endmember spectra update variants are described in Appendix \ref{ap:spectra}. 

\subsubsection{Sparsity Modulation Modalities}

The individual update steps of the BLUTH training procedure are linked together through a pattern of sparsity modulation, in which the sparsity regularization parameter ($\gamma$) is varied, in both the growth and fine-tuning stages. 
The exponential decay pattern proposed for DA-NMF, developed as the De-sparsify Modality (DeSM) here, is supplemented with 3 other modalities (Figs. \ref{fig:sparsity_phase_diagram} and \ref{fig:dynamic_penalties} and Table \ref{tab:SMmodalities}). 

The equilibrate modality (EqM), in which the endmember and abundance matrices are updated with alternation minimization and a constant sparsity regularization level $\gamma$, for $n_{\text{runs}}$ steps before stopping, provides a stable basis for the training and is re-used as part of the Shake and De-sparsify modalities. 

\begin{table}
\caption{Sparsity Modulation Modalities}
\begin{tabular}{c|c|c}
     Modality & Core Pattern & Termination \\   
     \hline
     Equilibrate & Constant $\gamma$ & After $n_{\text{runs}}$ \\
     Sparsify & Increase $\gamma$ each step for $n_{\text{runs}}$ & PPP Setpoint \\
     Shake & Turn $\gamma$ on every other step & Objective increases  \\
     De-sparsify  & Decay of negative $\gamma$ & After $n_{\text{runs}}$
\end{tabular}
\label{tab:SMmodalities}

\end{table}

The Sparsify modality (SpM) gradually increases the sparsity the abundances to the level specified by the PPP setpoint, a hyperparameter set at the start of the training, see Appendix \ref{ap:hyperparameters}. 
Because the application of sparsity tends to lock the abundances into local minima, the sparsity regularization periodically vanishes to ensure that the abundances do not get trapped in a local minima.
SpM increases $\gamma$ with a fixed step size for $n_{\text{runs}}$ updates.
If the PPP is then above the setpoint, the SpM concludes and the training algorithm proceeds to the next modality. 
However, if the PPP is still below the setpoint, the $\gamma$ step size is increased, and $\gamma$ is increased again from 0. 
The process is until the PPP setpoint is reached.

SpM utilizes the hierarchical structure of the BLUTH. 
The modulation begins by adjusting $\gamma$ only for the highest level of the network.
After the PPP setpoint is reached for the highest level of the network, SpM begins to modulate $\gamma$ for the two highest levels.
The process is repeated until all levels reach the PPP setpoint. 
Unlike prior group sparsity penalties, which sought to control both inter- and intragroup sparsity \cite{drumetz_hyperspectral_2019}, SpM does not explicitly distinguish between groups, but instead sequentially imparts sparsity to each level of the hierarchy.

The Shake Modality (ShM) forces sparsity into the network without increasing the data-consistency term of the objective function. 
The ShM consists of two stages: (i) first, the EqM is run once with $\gamma = 0$ and then (ii) a positive $\gamma$ is applied to the objective function on every other step, with $\gamma=0$ during the alternate steps for $n_{\text{runs}}$. 
The value and standard deviation of the data-consistency term of the objective function are estimated in the first run of (i).
After stage (ii) is applied, stage (i) is run again, and the system relaxes to a local minimum while the data-consistency term is estimated again. 
Then, if the minimum value of the objective is less than the estimate from the initial run of stage (i), stage (ii) is run again. 
The alternation between stages (i) and (ii) repeats until the minimum value of the objective function after stage (ii) is greater than the initial estimate, then the ShM ends.

The De-sparsify modality (DeSM) decreases the sparsity of the system, to escape local minimum. 
The DeSM is used in both the growth stage and the fine tuning stage, but appears in the former in an abbreviated form.  
In the fine-tuning stage, DeSM decreases the sparsity of the system by applying a negative value of $\gamma$ in a dynamic, exponentially decreasing pattern, akin to DA-NMF (Appendix \ref{ap:annealing}). 
The DeSM modality is formed by joining the EqM and ShM modalities. 
For each value of $\gamma$, the EqM is run once, followed by the ShM to prevent the unique spatial patterns of small endmembers from vanishing. 

Because of the time requirements of DeSM ($O(n_{\text{runs}}^{2})$ at a minimum), an abbreviated version is used in the growth stage, called DeSM-g, which uses linear support vector machines (SVM) to accelerate training \cite{chang_libsvm_2011}. 
At each node with descendants, the lowest-level endmember spectra of the positive and negative branches are used as pseudolabels for the SVM training. 
Because PPA is used for spectral updates during the growth stage, each labeled spectrum corresponds to a pixel in the data set. 
Since these spectra are already linearly separated by the BLUTH network, each endmember is guaranteed to have at least 1 pure pixel as partitioned by the SVM. 
However, since SVMs also maximize the margin between the labelled data, they maximize the number of mixed pixels while retaining the intricate spatial patterns. 
The weights from the SVMs are directly moved to update the $\mathbf{w}_{\zeta}$.
While DeSM-g works well in the growth stage, the use of AA spectral updates, rather than PPA, in the fine-tuning stage eliminates its pure pixel guarantee. 

\subsubsection{Sparsity Modulation Unmixing Growth (SMUG)}

The BLUTH network is grown in a greedy, exhaustive manner. 
When there are $p$ endmembers present, each endmember is split into 2 more in a separate network and stabilized. 
Then the new endmember split which decreases the objective function the most is retained.
The process is repeated until the stopping criterion, here the maximum number of endmembers, is met. 

The SMUG algorithm addresses one central challenge: new endmembers are initialized most clearly when the abundances are very sparse, but the estimates of how a new endmember will affect the objective function become more accurate after the system relaxes. 
The overall strategy of SMUG is then to (i) sparsify the abundances of the original $p$ endmembers (ii) split each endmember in a copy of the network, (iii) each copy is de-sparsified, (iv) the abundances equilibrate for each (v) select which copy to retain before restarting at (i) if the stopping criterion is not met. 
Each of the stages (i-iv) operates by applying the modalities listed above  (Fig. \ref{fig:hUH_process}). 
During (i) the SpM is applied to the entire network.

In stage (ii) the $\zeta$th endmember is split into two new endmembers in a copy of the original network. 
First, the pixels with $a_{\zeta}=1$ are split into two clusters and $\mathbf{w}_{\zeta}$ is initialized according to equation \ref{eq:lambda}. 
Second, EqM is applied to the pure pixels repeatedly, updating only the $\zeta$th endmember and its two descendants, until both have at least one pure pixel. 
Third, the rest of the nodes are then included and EqM is run once more. 
Finally, SpM is run on all the nodes, so that all nodes are at a similar level of sparsity. 
The initialization process is repeated for each split endmember. 

The desparsify stage (iii) applies DeSM-q to the nodes, so that they become less sparse but retain at least one pure pixel, before starting the relaxation stage (iv). 
Then the EqM is run twice with $\gamma=0$, once with only abundance matrix updates, and then with both abundance and spectra matrix updates. 
Next, the ShM is run to prevent the system from becoming trapped in insufficiently-sparse local minima. 
After that, the EqM is run twice more.

The value of the lowest level of the objective functions for each copy is then compared (v). The copy with the smallest objective function becomes the new base BLUTH. If the network has the desired number of endmembers is reached, then the training enters the fine-tuning stage, otherwise the sequence is repeated from step (i). 

\subsubsection{Fine-tuning stage}

In the fine-tuning stage, the abundances and endmember spectra are updated, but overall topology of the network is maintained (Fig. \ref{fig:hUH_process}b). 
In the transition from SMUG to fine-tuning, the SpM is applied to sparsify the whole network.
In the first part of the fine-tuning stage, the DeSM extracts the network from overly sparse local minima.
The $\gamma<0$ sections allow the network to explore more paths of relaxation, while ShM prevents small endmembers from vanishing. 

The final relaxation stage is designed to allow the system to gently come to rest in a minimum of the objective function which is reasonably sparse.
In this stage, only the lowest-level terms are considered in the objective function and all updates use the larger batch size. 
The final stage is run twice, once continuing with PPA updates to the abundance matrix and once with AA updates to the abundance matrix, so that the spectral matrix update techniques can be compared. 
After the stage is complete, the network is saved. 

\section{Experiments}

\subsection{Evaluation scenes}

Both remote sensing images and laboratory-acquired scenes are used for testing. 

Six sets of manual labels are used for the remote scenes, including the Samson, Jasper Ridge, APEX, Urban, and Washington DC scenes, with two sets of labels for the Urban scene. These scenes were compiled for testing in \cite{zouaoui_entropic_2023}, and were recorded by a variety of hyperspectral cameras over the visible and shortwave infrared portion of the electromagnetic spectrum. The labels were applied to these scenes manually in a process described in \cite{zhu_hyperspectral_2017}. The labels do not indicate a {\it ground truth}, but rather how a human analyst unmixed the datasets. 

Two scenes recorded in a laboratory complement the remote scenes \cite{cui_realistic_2023} with labels that function as {\it ground truth} because the proportion of each material in each pixel could be determined precisely. However, many of the effects which complicate remote sensing scenes (atmosphere, jitter, spectral variability) are either absent or much reduced because of the laboratory setting. Therefore, these scenes give a clear estimate of how an algorithm will perform in laboratory conditions, but not necessarily of how they will perform in the field. 

The two sets of scenes complement each other. The remote sensing scenes contain all the artifacts and complications of real data but have specious labels, while the laboratory scenes have reliable labels but lack the complications of remote sensing. Together, they provide a more comprehensive perspective on the performance of different unmixing algorithms. 

One ambition in the development of BLUTH was the unmixing of water masses. In the absence of test datasets with ground truth, a scene with a Coccolithophore bloom was unmixed. The goal of the test was to see if the BLUTH could separate the different water masses from each other according to their spectral signatures, as well as separate the water from clouds and land. Hyperspectral images of the bloom both from the large and extremely precise PACE satellite as well as the HYPSO-2 cubesat were unmixed \cite{grotte_ocean_2021,dierssen_synergies_2023}. 
Both images show the mouth of Oslofjord meeting the Skagerrak on May 28th, 2025. With PACE the L2 Near Real-Time Bottom-of-atmosphere reflectance is used. A $200\times 150$~pixel subscene is selected with 122 bands (Fig. \ref{fig:oceancolorhierarchies}b). The HYPSO scene is reduced to a $200\times 200$~pixel subscene with 105 of the original 120 bands retained. The first 8 bands, 5 bands around the 760~nm O$_{2}$ absorption line, and the last 2 bands are excluded from the analysis. A columnwise, smoothed dark pixel subtraction is applied to the L1D top-of-atmosphere reflectance HYPSO-2 scene as an approximate atmospheric correction \cite{chavez_improved_1988}.
 
\subsection{The evaluated techniques}

The experiments include unmixing methods that rely on different aspects of the BLUTH architecture: implicit endmember spectral constraints (EDAA, DAAA, SAPPA) \cite{zouaoui_entropic_2023}, hierarchical structure (H2NMF, DC-DNMF)\cite{gillis_hierarchical_2014,de_handschutter_consistent_2023}, and dynamic optimization procedures (DAAA, SAPPA)\cite{peng_nonnegative_2012} (Table \ref{tab:methods}).  These are compared against three unmixing techniques which achieved the state-of-the-art performance in the recent HySUPP unmixing comparison: MiSiCNet \cite{rasti_misicnet_2022}, MSNet\cite{yu_multi-stage_2022}, and Nonnegative Matrix Factorization-Quadratic Minimum Volume (NMF-QMV) \cite{zhuang_regularization_2019,rasti_image_2024}. The BLUTH architecture itself is trained in both pure pixel analysis (PPA) and archetypal analysis (AA) forms. To test the robustness of the BLUTH algorithm against errors in the number of endmembers, both variants are also trained with two additional endmembers. 

\vspace{5pt}

\begin{table}[h]
\begin{tabular}{l||c|c|c|c}
Method & Hierarchy & ${S_m}$  & Dynamic & HySUPP  \\
& & Constraint &  & SoTA\\
\hline
MiSiCNet \cite{rasti_misicnet_2022} & & & & $\checkmark$ \\
MSNet \cite{yu_multi-stage_2022} & & & &$\checkmark$ \\
NMF-QMV \cite{zhuang_regularization_2019} & & & &$\checkmark$ \\
H2NMF \cite{gillis_hierarchical_2014} & $\checkmark$ & $\approx$PPA & & \\
DC-DNMF \cite{de_handschutter_consistent_2023} & $\checkmark$ & & & \\
EDAA \cite{zouaoui_entropic_2023}& & AA & & $\checkmark$ \\
DAAA \cite{peng_nonnegative_2012}& & AA & $\checkmark$ & \\
SAPPA \cite{kirkpatrick_optimization_1983,peng_nonnegative_2012}& & PPA & $\checkmark$ & \\
BLUTH-AA &$\checkmark$&AA&$\checkmark$&\\
BLUTH-PPA &$\checkmark$&PPA&$\checkmark$&\\
\hline
\end{tabular}
\caption{The evaluated unmixing methods}
\label{tab:methods}
\end{table}

\begin{figure*}
\centering
\includesvg[width=0.95\textwidth]{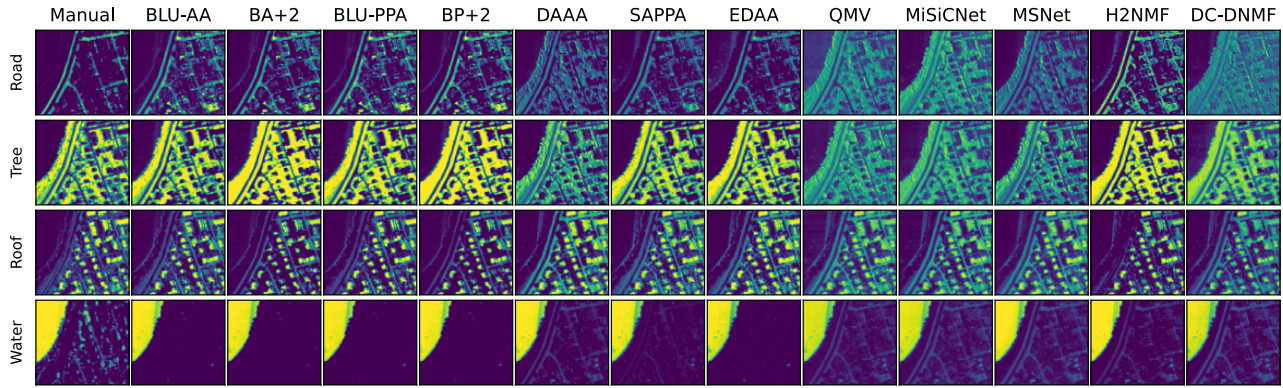}
\caption{The APEX scene unmixed into Road, Tree, Roof and Water endmembers.}
\label{fig:Apex_spatial}
\end{figure*}

\subsection{Evaluation criteria}

The abundances and endmember spectra are evaluated separately in terms of their similarity to a manually-processed endmembers. 
The accuracy of the reconstructed endmember spectra is evaluated through the spectral angle with respect to the manually-processed spectral angle, expressed in degrees:
\begin{equation}
    \Phi(\eta,\zeta) = \cos^{-1}\Bigg( \frac{\bm{s}_{\eta}^{T}\bm{s}_{\zeta}}{\|\bm{s}_{\eta}\|_{2}\|\bm{s}_{\zeta}\|_{2}}\Bigg),
    \label{eq:spectral_angle}
\end{equation}
where $\bm{s}_{\eta}$ and $\bm{s}_{\zeta}$ are the estimated and manually processed spectra, respectively. 
The spectral angle approaches zero as the estimated spectra becomes more similar to the manually-processed spectra, but tends towards 180 degrees as they become more dissimilar. 

The Intersection over Union (IoU) metric is often used to characterize the performance of tasks such as clustering and classification. 
The ASC allows IoU to be generalized to unmixing:
\begin{equation}
    \text{IoU}(\eta,\zeta) = \frac{\sum_{n}^{N}\min(a_{\eta n},a_{\zeta n})}{\sum_{n}^{N}\max(a_{\eta n},a_{\zeta n})},
    \label{eq:IoU}
\end{equation}
where $a_{\zeta n}$ and $a_{\eta n}$ refer to the estimated and labeled endmember abundance at pixel $n$.
Unlike other metrics used to characterize the abundance estimation portion of unmixing, such as the root mean square error, IoU is bounded between 0 and 1, which permits comparison between endmembers.
Other generalizations consistent with the clustering definition of IoU could be defined, but equation \ref{eq:IoU} is used because of its simplicity. 

\section{Results}

The unmixing techniques generally achieve decent performance on most of the datasets (Tables \ref{tab:SA_test} and \ref{tab:IoU_tests}). A threshold of ten degrees above the minimum spectral angle for an endmember is used to evaluate whether a technique has found it. 

All the unmixing techniques visually appear to find the three correct endmembers in the Samson scene (Figs. \ref{fig:Samson_spatial} and \ref{fig:Samson_spectral}). Algorithms based on AA updates of the spectra perform the best on all endmembers. 
The two BLUTH networks with additional endmembers show only slightly degraded performance, with the most significant degredation occuring for {\it water}.

Half of the techniques find the four endmembers on the Jasper Ridge scene (Fig \ref{fig:Jasper_spatial}), while the other half either lack the {\it road} endmember or the {\it dirt} endmember. The best performance is again achieved by the AA-based algorithms, except DAAA, while the techniques NMF-QMV, MiSiCNet, MSNet, and DC-DNMF all predict inadequately sparse endmember abundances.
The two BLUTH networks with additional endmembers do detect all endmembers in the manual labels, but exhibit excessive sparsity. 

Seven of the techniques find the proper endmembers in the Apex scene (Fig \ref{fig:Apex_spatial}). The most common failure mode is to miss the {\it road} endmember and instead split the {\it tree} endmember into two, as exemplified by DAAA, NMF-QMV, MiSiCNet, MSNet, and DC-DNMF. 
As before, the techniques which rely on AA perform the best spectral estimation, but each endmember spectra is estimated best with a different technique. 
While EDAA achieves the highest IoU on three of four endmembers, the second-highest IoU is achieved by a BLUTH technique. 
The two BLUTH networks with additional endmembers show performance comparable to those with the original number of endmembers, even achieving the two lowest {\it roof} spectral angles. 

The Urban scene is processed with two different sets of labels: one with 4 endmembers and one with 6. 
When only 4 endmembers are considered, the {\it roof} endmember is missed by 7 of the techniques (according to the spectral angle threshold) (Fig. \ref{fig:U4_spatial}) because of the inadequate sparsity failure mode. However, when 6 endmembers are considered, only EDAA finds all 6 endmembers (Fig. \ref{fig:U6_spatial}). Most of the techniques locate shadows rather than the {\it metal} endmember. 
On these scenes, the BLUTH networks with two extra endmembers produce comparable results in terms of correspondence with the manual labels, but they also exhibit more sparsity, particularly in 4-endmember case. 

No technique clearly unmixes all 6 endmembers in the Washington, DC scene when using the designated number of endmembers, although the two BLUTH networks with extra endmembers do (Fig. \ref{fig:WDC_spatial}). Only H2NMF, DAAA, DC-DNMF, and SAPPA locate more than half of the {\it grass} abundance, but they all lack the {\it roof} endmember. Only BLUTH-AA and NMF-QMV, of the techniques using the designated number of endmembers, find the {\it roof} endmember. 

\begin{figure*}
\centering
\includesvg[width=0.95\textwidth]{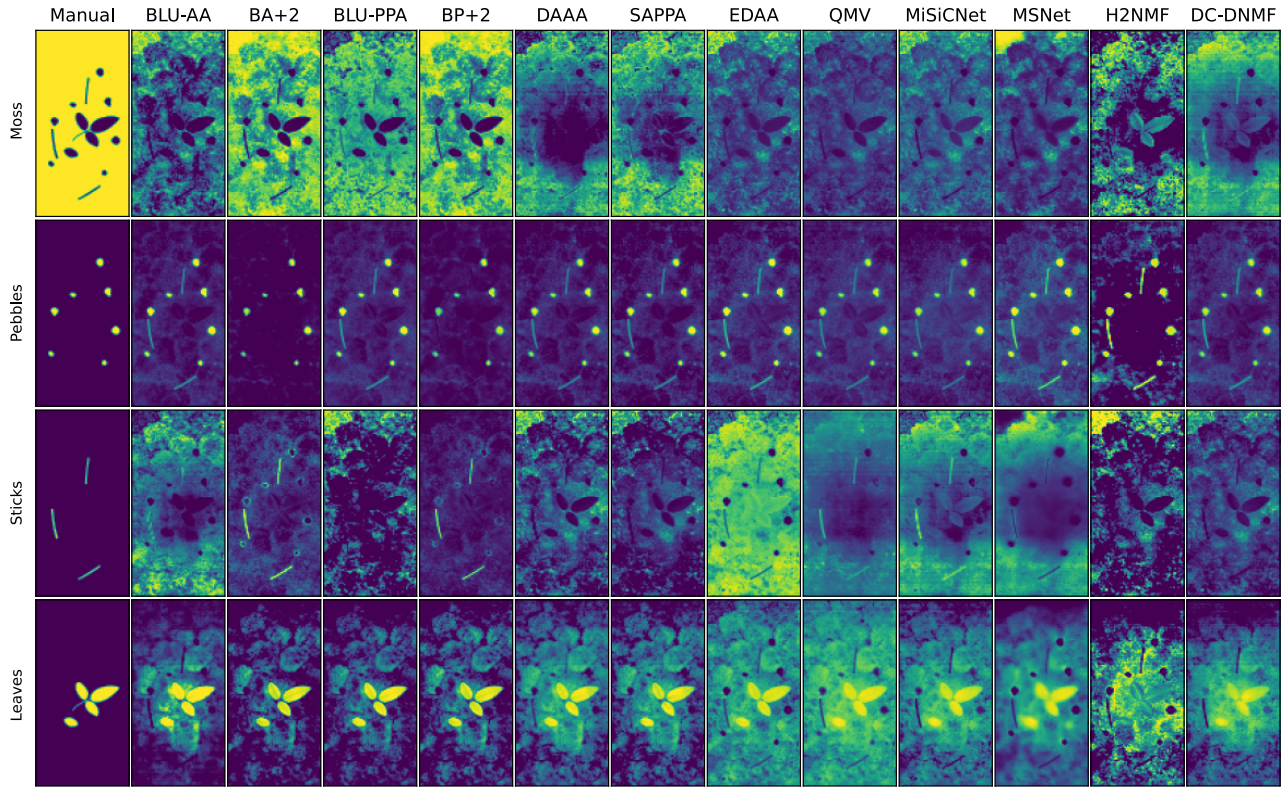}
\caption{The abundances of the RMMS-Simple Mixing Scene unmixed into 4 endmembers.}
\label{fig:SMS_spatial}
\end{figure*}

The Simple Mixing Scene of the Realistic Mixing Miniature Scenes causes all of the techniques to produce inadequately sparse endmember abundances (Fig \ref{fig:SMS_spatial}). However, the spectral angle calculation shows that some of the techniques find all 4 endmembers despite an excess of mixed pixels. The {\it moss} endmember achieves an IoU above 0.5 only if unmixed by BLUTH with excess endmembers or BLUTH-PPA. No techniques unmix {\it Pebbles} to an IoU above 0.5, but only NMF-QMV and MSNet fall below 0.1. 
Even though the IoU values for the {\it sticks} endmember are all very small because it is present in such a small number of pixels, there is clear variation between the techniques. The techniques BLUTH-AA, BLUTH+2, and MiSiCNet all detect it. The BLUTH+2 networks detect it most clearly, with the AA variant achieving a spectral angle below 5 degrees, less than half the spectral angle of any of the techniques using the designated number of endmembers.

The Complex Mixing Scene of the Realistic Mixing Miniature Scenes also causes all of the techniques to produce inadequately sparse endmember abundances (Fig \ref{fig:CMS_spatial}). All of the endmembers of the scene were missed by at least one unmixing technique. The IoU of {\it moss} exceeds 0.5 for all of the hierarchical techniques, but none of the others. The {\it pebbles} endmember IoU only exceeds 0.1 for the hierarchical techniques, excluding DC-DNMF. The {\it sticks} endmember was particularly challenging, and its IoU only exceeded 0.02 for the BLUTH+2 networks. More consistent behavior was observed in the detection of the {\it true vegetation} class, as all techniques except EDAA, MiSiCNet, and MSNet achieved IoUs above 0.2 and spectral angles below 4 degrees. All techniques except MiSiCNet and DC-DNMF achieved IoUs above 0.5 and spectral angles below 5 degrees for the {\it false vegetation} endmember. 

\begin{figure}
\centering
\includegraphics[width=0.5\textwidth]{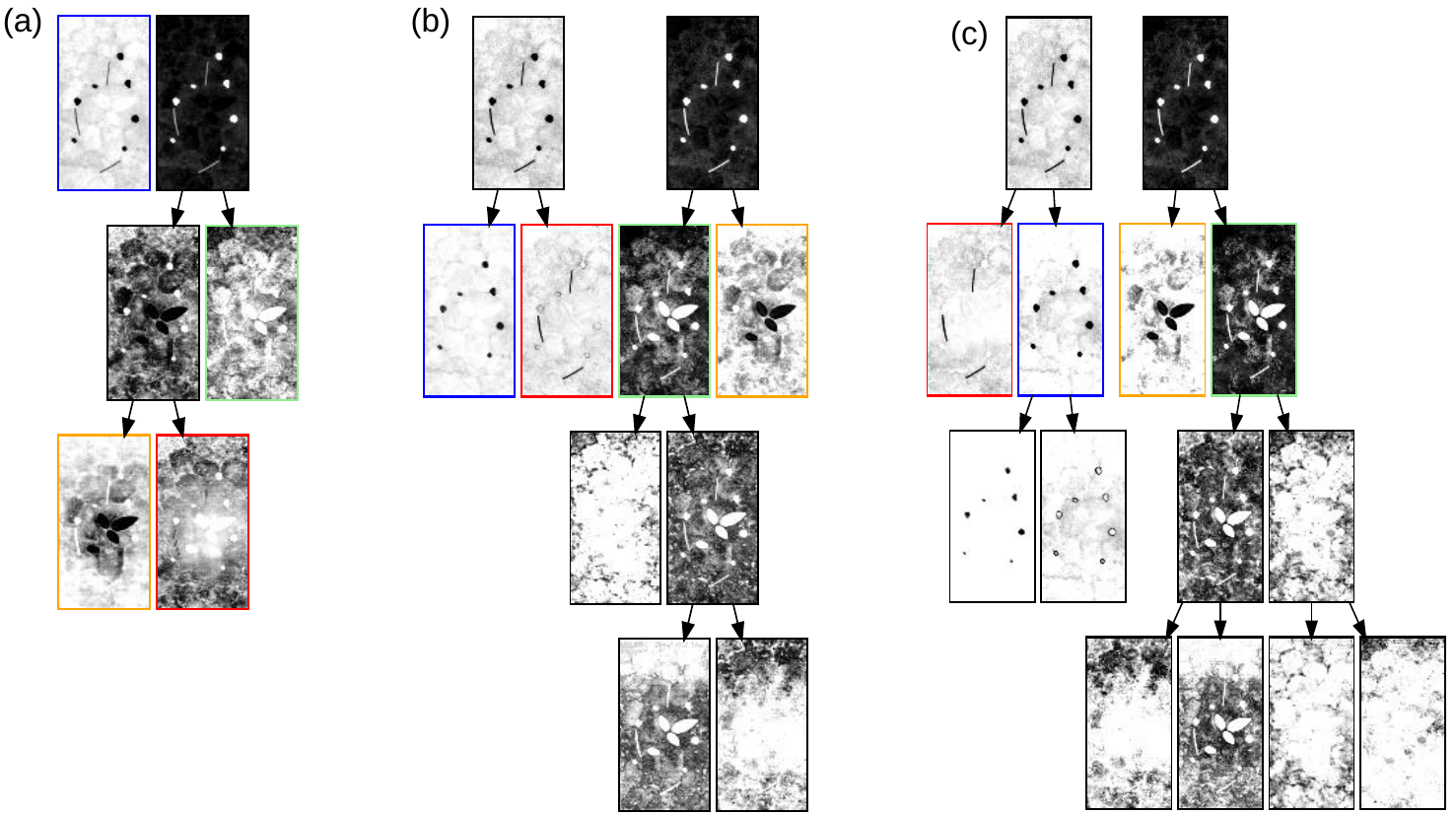}
\caption{The Simple Mixing Scene unmixed with 4 (a), 6 (b) or 8 (c) endmembers with BLUTH-AA. The highlighted abundances are determined to correspond to the endmembers manually-labelled endmembers as moss (green), pebbles (blue), sticks (red), and leaves (orange).}
\label{fig:SMS_hierarchies}
\end{figure}

\begin{figure}[t]
\centering
\includesvg[width=0.47\textwidth]{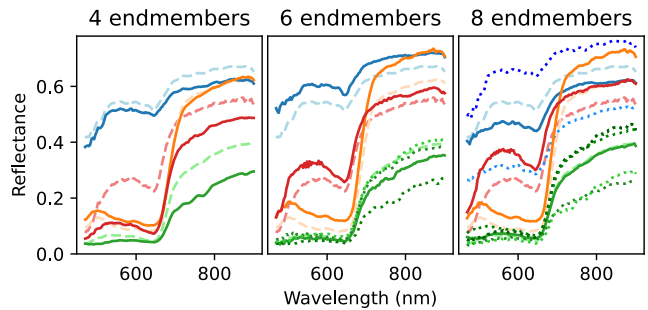}
\caption{The endmember spectra in the RMMS-Simple Mixing Scene. 
The spectra of the 4 selected endmembers (solid lines) are compared to the manually-labeled (dashed lines) moss (green), pebbles (blue), sticks (red), and leaves (orange). 
Spectra from descendants of the selected endmembers are shown as dotted lines, with dark$\rightarrow$light representing left$\rightarrow$right in Fig. \ref{fig:SMS_hierarchies}. }
\label{fig:SMS_hierarchy_spectra}
\end{figure}

The performance of the BLUTHs with extra endmembers motivated an investigation into how their architecture grew as the number of endmembers increased on the Simple Mixing Scene (Fig. \ref{fig:SMS_hierarchies}). BLUTH networks were trained with 4, 6, and 8 endmembers. With four endmembers, the endmembers occur at different levels of the hierarchy. The highest level shows a clear split between vegetation and non-vegetation, but the vegetation is partitioned into three endmembers, and the non-vegetation is not partitioned at all. The spectra of the recovered endmembers show what has happened in more detail (Fig. \ref{fig:SMS_hierarchy_spectra}). The spectra of the {\it leaves} is recovered fairly well, but the spectra of the endmembers which are identified as the {\it pebbles}, {\it sticks} and {\it moss} are all notably darker than the endmembers in the manual labels. The spectrum identified as {\it moss} is in fact much darker than any spectrum in the manual labels. In essence, shadows in the image have been assigned to the {\it moss} endmember and the {\it sticks} endmember is a combination of the {\it sticks} and {\it moss} spectra in the manual labels. Overall, the abundances succumb to the inadequate abundance sparsity failure mode. BLUTH-PPA with 4 endmembers shows similar results, but with the final identification of {\it moss} and {\it sticks} switched because of small changes in their abundances. 

The abundances of the selected endmembers become more sparse as the number of endmembers included in the network increases, although the sparsity at the highest level decreases. The endmembers included in the manual labels are all at the second level for networks with both 6 and 8 endmembers. With 6, the endmember selected as {\it moss} is split into two endmembers at the third level, which roughly correspond to {\it moss} itself and shadows, according to the spectra in Fig. \ref{fig:SMS_hierarchy_spectra}b. The spectrum corresponding to {\it moss} is then split into two more endmembers, which are positioned towards the center and the edge of the scene. In addition, with 6 members, the {\it pebble}, {\it stick}, and {\it leaves} spectra are brighter than indicated in the manual labels, while the {\it moss} is still darker. The basic structure of the 6-endmember hierarchy remains the same when two more endmembers are added. The new endmembers simply split the {\it pebble} and shadow endmembers. The abundances of the endmembers identified with those from the manual labels do not change noticeably. However, the spectra in Fig \ref{fig:SMS_hierarchy_spectra}c show that while {\it moss} is close to the spectrum in the manual labels, both {\it sticks} and {\it leaves} are brighter than those in the labels, while {\it pebbles} is darker. For 6 and 8 endmembers, BLUTH-AA and BLUTH-PPA show similar results. 

\begin{figure*}[t]
\centering
\includegraphics[width=0.95\textwidth]{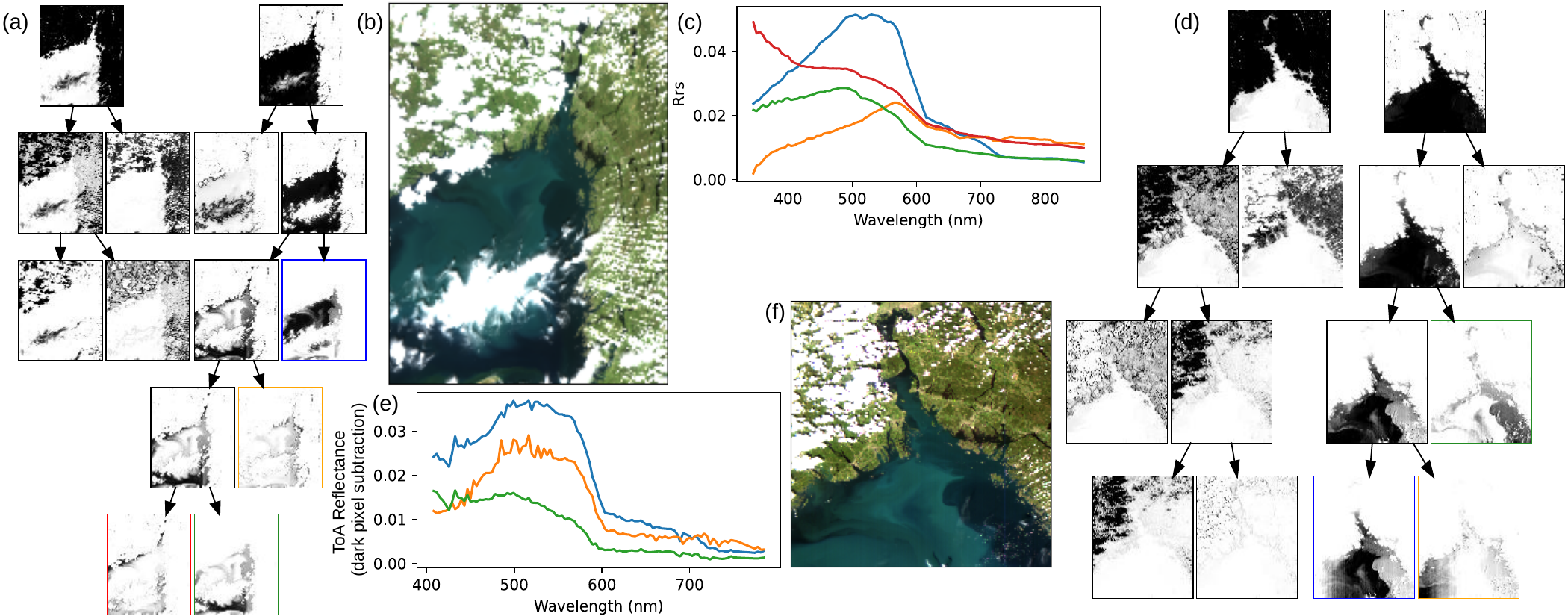}
\caption{Unmixing hierarchies can separate water masses both in data collected by large hyperspectral missions such as the PACE satellite (a,b,c) as well as from smaller hyperspectral cubesats, such as HYPSO-2 (d,e,f). The images show a coccolithophore in the Skagerrak on May 28, 2025. The spatial maps of the aquatic endmember abundances are highlighted in the colors in which their spectra are plotted in the next figure.}
\label{fig:oceancolorhierarchies}
\end{figure*}

In the final tests, the BLUTH unmixing of the Skagerrak scenes is able to clearly resolve different portions of the water (Fig. \ref{fig:oceancolorhierarchies}). The split between land and water occurs at the first level of the hierarchy, and clouds are split off just below that, for both the PACE and HYPSO images. Spectral features relating to the coccolithophorid bloom (blue) as well as CDOM-dominated river run-off (yellow) can be clearly distinguished in both sets of unmixed spectra, although PACE is much less noisy. Weights were added to standardize the contribution of each pixel to the objective function so that the land pixels would not contribute a disproportionate amount, based on preliminary tests.

\section{Discussion}

Overall, the above experiments show that BLUTHs have the strongest performance of any of the techniques in several aspects, but are also competitive in general. To aid in the interpretation of the IoU and spectral angle metrics, which vary quite a bit between endmembers, the number of endmembers missed by each unmixing technique in each scene is tallied in Table \ref{tab:Misses}. Although the BLUTH-based techniques generally show a slightly lower IoU on the remote scenes than EDAA, they show comparable spectral angles for most endmembers. On the laboratory scenes, the BLUTH-based techniques record the top two IoU performances for every endmember, and include the smallest spectral angle for 4/9 endmembers, with the others split between the annealing techniques and H2NMF. Increasing the depth of the BLUTH networks by 2 endmembers only harmed the performance on one scene (Jasper Ridge) and improved it on the laboratory scenes. BLUTH-AA seems slightly more robust than BLUTH-PPA, but the difference is small relative to the difference with other techniques (Fig. \ref{fig:J_unmixing_consistency}). Moreover, when the BLUTH algorithms fail, such as when shadows are found instead of {\it metal} on the Urban6 scene, it does not necessarily imply worse behavior in the ocean color use case. It merely shows that BLUTHs are sensitive to dark pixels, which is not terribly prohibitive to partitioning (relatively dark) water masses.

The small file size required to store the output of a BLUTH was noted during testing to be another benefit. Most of the unmixing techniques require storing a dictionary of size $PB$ as well as a map of size $PN$. However, BLUTHs store the function for mapping pixels to abundances explicitly, instead of storing the abundance per pixel. Therefore, they require $PB$ to store the spectra of the endmember nodes, but only require $(P-1)(2B+1)$ to store the function for partitioning abundances. This leads to significant reduction of storage needs. For example, the Urban with 6 endmembers requires 2.3 MB with the other techniques but a BLUTH only requires 10.4 kB with 4-byte floats. 

Among the other tested techniques, several other patterns are apparent. The techniques with constrained endmember selection (AA or PPA) show much better performance overall than those without. This effect is quite strong; every technique that uses AA or PPA misses fewer endmembers than every technique that does not impose those constraints. One complication here is that all of those techniques also incorporate a strategy to deal with the nonconvex objective function: SMUG (BLUTHs), annealing (DAAA and SAPPA), or multiple runs (EDAA). 

The effects of hierarchy are more ambiguous. While the BLUTHs themselves perform well, and H2NMF records comparable performance, DC-DNMF performs relatively poorly. H2NMF implicitly adheres to the HASC during its clustering stage (although not explicitly stated in \cite{gillis_hierarchical_2014}), and BLUTHs adhere to it at all stages, while DC-DNMF only adheres to the ordinary ASC. DC-DNMF's performance could be due to the absence of HASC. 

A notable difference exists between the remote scenes and the laboratory scenes. For example, EDAA achieves top-2 IoU performance on over half of the remote endmembers, but on none of the laboratory endmembers, while BLUTHs perform better on the laboratory scenes than on the remote sensing scenes. 
One immediate visual difference is that the laboratory endmember labels are much sparser than the labels of the remote scenes. However, there may be other aspects of the images which contribute to the behavior gap. Inadequatly sparse abundances seems to be the most common failure mode, except for H2NMF.

\begin{table}[]
    \centering
    \begin{tabular}{l||c|c|c|c|c|c|c|c||r}
Technique &Sa&J&A&U4&U6&W&SM&CM&Total\\
\hline
BLU-AA  &0&0&0&1&1&2&0&0&4\\
BA+2    &0&0&0&0&1&0&0&0&1\\
BLU-PPA &0&0&0&1&1&2&1&0&5\\
BP+2    &0&0&0&0&1&0&0&0&1\\
DAAA	&0&1&1&1&2&3&1&0&9\\
SAPPA   &0&0&0&1&1&1&1&0&4\\
EDAA    &0&0&0&0&2&1&1&3&7\\
NMF-QMV &1&4&2&1&5&4&3&2&22\\
MiSiCNet&1&3&2&3&6&6&1&4&26\\
MSNet   &0&1&1&0&2&3&1&2&10 \\
H2NMF   &0&1&0&0&1&1&2&1&6\\
DC-DNMF &0&2&2&2&4&5&1&3&19
    \end{tabular}
    \caption{Number of missed endmembers on each scene, defined as a spectral angle more than 10 degrees above the minimum.}
    \label{tab:Misses}
\end{table}

\section{Conclusion}

The hierarchical abundance sum constraint, introduced above, provides the foundation of the BLUTH unmixing. The SMUG algorithm for growing the BLUTH networks can achieve stable training and accurate endmember estimates. The BLUTHs themselves, although they are built on a simple architecture, give performances which exceed or are comparable to the state-of-the-art unmixing techniques on the datasets evaluated. Tests on unlabeled images over water bodies suggest that the BLUTH networks are appropriate for partitioning hyperspectral ocean color images. 

In the course of these tests a few relatively neglected unmixing techniques, including H2NMF and deterministic annealing, which were included because of technical similarities with the proposed BLUTHs, were shown to have behavior competitive with state-of-the-art techniques. This suggests that other algorithms with surprisingly good performance may reside in the older unmixing literature, and that pre-2015 methods should still be included in comparisons of unmixing performance. On that note, although BLUTHs are technically deep neural networks (multi-layered networks with neurons and activation functions), they functionally share little with deep learning: the outputs are interpretable at every level of the hierarchy, the weights of each node can be interpreted independently, and BLUTHs can be trained on small datasets. 

Accelerating the training speed of the BLUTH networks is the greatest challenge for applying them to large datasets. Although duration was not a primary concern here, BLUTHs required about 3 hours to train on the Samson scene, and over a day to train on the Washington, DC scene. These are still relatively small scenes, with at most $\approx10^{5}$ pixels. A single scene from the HYPSO constellation contains $6\times$ as many pixels. The HYPSO database of over 3000 images contains $>10^{9}$ pixels, and it is a fairly small database compared to larger, monolithic satellites currently in orbit. The PACE satellite, which collects data continually, has many times that. It is hoped that optimization techniques borrowed from other unmixing algorithms, particularly EDAA, might help to bridge those four orders of magnitude, so that the entire database can be jointly unmixed. The SMUG algorithm, since it requires $p$ copies of the network to be re-trained each time a new endmember is added, dominates the computational time requirement. 

There are also many small changes to the networks which the above results suggest could bring significant performance improvements. 
First, the layerwise objective function weighting could be explored in more detail, as we expect that it could have a significant influence on the output. The Pure Pixel Proportion setpoint and batch sizes did not seem to have a large effect on the output, but could be explored further. In addition, batching is the only way that randomness enters the SMUG algorithm, so it could be used to modulate randomness. 

Numerous new unmixing possibilities are opened up by the HASC beyond BLUTHs. For example, one simple generalization could be to allow a node to split into more than two new endmembers at a time. This would aid in cases where there is ambiguity regarding 2-way splits. For example, should clouds be grouped with water or land? Moreover, the constraint could be incorporated into older Deep NMF techniques, such as DC-DNMF. Although SMUG was developed for growing a network, the framework of sparsity modulation could bring general benefits towards exploring the solution space of the nonconvex unmixing problem, perhaps combined with the method of multiple solutions used by EDAA.

\section*{Acknowledgment}

This work was supported in part by NO Grants 2014 – 2021, under Project ELO-Hyp, contract no. 24/2020, in part by the Research Council of Norway and Industry Partners through SoS (337495) and HYPSCI (325961), and in part by the European Union under the Horizon Europe Programme, Grant Agreement No. 101082004 (DiverSea). Views and opinions expressed are however those of the author(s) only and do not necessarily reflect those of the European Union or European Research Executive Agency (REA). Neither the European Union nor the granting authority can be held responsible for them.

We would like to thank the scientists and engineers who shared their data and software. The HYPSO-2 image can be found at \url{https://hypso.space/dataportal/}. The PACE image can be found at \url{https://pace.oceansciences.org/access_pace_data.htm}. The Remote Sensing test scenes can be downloaded from \url{https://github.com/inria-thoth/EDAA}. The Realistic Miniature Mixing Scenes can be downloaded at \url{https://data.mendeley.com/datasets/6bg6342xg9/1}. The EDAA, NMF-QMV, and MiSiCNet algorithms are part of the HySUPP package \url{https://github.com/inria-thoth/HySUPP}. The H2NMF and DC-DNMF algorithms can be downloaded from \url{https://sites.google.com/site/nicolasgillis/code}. 

Code to train BLUTHs is available at \url{https://git.ntnu.no/NTNU-SmallSatLab/BLUTH}. We acknowledge the use of Numpy and Scikit-learn (as a python interface to libsvm) in the BLUTHs themselves, as well as pyTables for saving BLUTH weights, Scipy for endmember/label identification, and graphviz and matplotlib for visualization \cite{harris_array_2020,pedregosa_scikit-learn_2011,chang_libsvm_2011,alted_pytables_2003,virtanen_scipy_2020,farin_graphviz_2004,barrett_matplotlibportable_2005}.

\nocite{metropolis_equation_1953}
\bibliographystyle{ieee_firstnames}
\bibliography{main}

\clearpage

\setcounter{figure}{0}
\setcounter{table}{0}
\setcounter{section}{0}
\setcounter{equation}{0}
\begin{appendices}
\fancyhead{} 

\renewcommand{\appendixname}{}
\appendix
\renewcommand{\thetable}{S-\Roman{table}}
\renewcommand{\thefigure}{S\arabic{figure}}
\renewcommand{\thesection}{S\arabic{section}}
\renewcommand{\theequation}{S\arabic{equation}}

\section*{Supplemental Information}

\subsection{Normalization}
\label{ap:normalization}

The normalization of the pixels mitigates the effect of the scaling degree of freedom which plagues non-negative matrix factorization and related techniques.
L1-normalization is common in optical water type analysis, in which it is called spectral curve integral normalization \cite{spyrakos_optical_2018}.
On the other hand, L2-normalization is proposed in \cite{zouaoui_entropic_2023}.
Paired with the squared error in the objective function, the L2 norm gives equal weight to each pixel.
While normalization improves overall performance, it also prevents the endmember spectra from being directly interpreted as remote sensing reflectance themselves.
Because the normalized spectra lack information about their original magnitude, they cannot be inverted back into reflectance or radiance. 
Here, this trade-off is negotiated by partial L2-normalization. 
A partially L2-normalized spectra can be written as:
\begin{align}
\tilde{\bm{y}} = \frac{\bm{y}}{(\|\bm{y}\|_{2})^{1-\epsilon}},
\end{align}
where $\epsilon \in [0,1]$ determines the degree of normalization. 
If $\epsilon > 0$, then the endmember spectra are invertible.
More precisely,
\begin{equation}
    \bm{y} = \tilde{\bm{y}}\|\tilde{\bm{y}}\|_{2}^{(1-\epsilon)/\epsilon}.
\end{equation}

Within the context of the LMM, normalization allows the magnitude of the endmember spectra to vary with the pixel brightness to accommodate spectral variation. 
At pixel $x$, endmember spectra $s$ has an effective magnitude of $\|\bm{y}\|_{2}^{1-\epsilon}\|s\|_{2}^{\epsilon}$.
The choice of $\epsilon$ greatly effects unmixing performance, and the requirement of invertibility offers no guidelines for choosing $\epsilon$ except $\epsilon>0$. 
Scale independence implies that multiplying the data by a constant should not change algorithm performance, which suggests that $\epsilon$ should depend only on ratios of pixel brightness, such as $\|\bm{y}_{\max}\|_{2}/\|\bm{y}_{\min}\|_{2}$, rather than the pixel values themselves. 
If the luminosity range scales with the number of endmembers raised to some power, then
\begin{align}
p^{\nu} &=(\|\bm{y}_{\max}\|_{2}/\|\bm{y}_{\min}\|)^{\epsilon} \\
\implies \epsilon = &\frac{\nu \log(p)}{\log\bigg(\frac{\|\bm{y}_{\max}\|_{2}}{\|\bm{y}_{\min}\|_{2}}\bigg)}.
\label{eq:epsilon}
\end{align}
The value of $\nu$ is set to be $1/4$ throughout all experiments, based on initial tests on the Jasper Ridge scene  (Table \ref{tab:Jmetrics}). 
Then neither $\epsilon$ nor $\nu$ remain as variable hyperparameters.

\subsection{Hyperparameters}
\label{ap:hyperparameters}

There are 6 hyperparameters which must be set to train a BLUTH. 
In general, the only hyperparameters that must be varied between images are the standard and large batch sizes. 
The pure pixel proportion (PPP) setpoint is used in the SpM. In general, increasing the PPP setpoint improves accuracy but also greatly increases calculation time. In the experiments presented here, it is either set to 0.5, or 0.8 if there are classes which are much less abundant than the others. 
The scaling between layers is set such that $\mu_{m+1}=4\mu_{m}$ (equation \ref{eq:obj}) and $\nu=0.25$ for the partial normalization exponent, and $n_{\text{runs}}=10$ for all experiments. 

The methods from the HySUPP review and H2NMF used the hyperparameters suggested in their codebases. DC-DNMF was only adjusted from its default settings by imposing the abundance sum constraint. A few time constants and iteration counts for DAAA and SAPPA were tested, and not much variation was seen as long as both were sufficiently large.

\subsection{The competing unmixing techniques}
\label{ap:competing_techniques}

Deterministic Annealing Archetypal Analysis (DAAA) and Simulated Annealing Pure Pixel Analysis (SAPPA) were both developed from Deterministic Annealing Nonnegative Matrix Factorization \cite{peng_nonnegative_2012}. However, because the unavailability of that code and ambiguities regarding the implementation of the abundance sum constraint, the method was implemented again with some simple changes. The implementations used here are described in Appendix \ref{ap:annealing}. The $\nu=1/4$ normalization, used by the BLUTHs, is also used for these techniques.

Entropic Descent Archetypal Analysis (EDAA) \cite{zouaoui_entropic_2023} combines Archetypal Analysis with updates which are designed to be accelerated by GPUs. Because EDAA is extremely fast, it can be run for many different random initializations and hyperparameters. Because the unmixing objective function is nonconvex, these different runs typically stop at different minima. A combination of the reconstruction accuracy and the endmember spectral coherence is then used to select a particular run of unmixing. The original presentation of EDAA found extremely high performance in abundance estimation, surpassing deep learning techniques, but the later HySUPP analysis showed mediocre performance in reconstruction accuracy, while it still achieved the lowest spectral angle of the blind methods on one dataset. The L2-normalization specified in its original presentation is used here. The implementation presented in HySUPP is used for testing here \cite{rasti_image_2024}. 

MiSiCNet is a deep network which constrains its output to the simplex and incorporates both spatial and spectral information \cite{rasti_misicnet_2022}. It performs backpropagation in PyTorch on the unmixing objective function with a regularization term acting on the endmember geometry. It exhibited the best performance on two of the three blind unmixing datasets in the HySUPP comparison, and the best spectral on one of them \cite{rasti_image_2024}. 

MSNet is another deep convolutional neural network autoencoder that unmixes scenes at 3 different levels of spatial resolution \cite{yu_multi-stage_2022}. It exhibited the best performance on one of the three blind unmixing datasets in the HySUPP comparison \cite{rasti_image_2024}. 

Nonnegative Matrix Factorization-Quadratic Minimum Volume incorporates a sparsity penalty not on the abundances, but rather on the endmember spectra and \cite{zhuang_regularization_2019} provides a way to estimate the proper weight to apply to the regularization term. It provided the second best abundance estimation performance on all of the HySUPP blind unmixing tests, as well as the lowest spectral angle on one dataset \cite{rasti_image_2024}.

Hierarchical Clustering of an HSI based on Rank-Two NMF (H2NMF) \cite{gillis_hierarchical_2014} explicitly connects unmixing and clustering. It is based an a divisive hierarchical clustering algorithm which, like the BLUTH architecture, successively splits clusters into two. Once image is split into clusters, the endmember spectra are selected from the pixels of each cluster based the mean-removed spectral angle. The recovered spectra are used to estimate the abundances. Here, H2NMF was applied to L2-normalized data. The original H2NMF implementation in MatLab is used. Notably, the H2NMF algorithm implicitly satisfies both the HASC, because of its reliance on clustering, and PPA, because of how the endmember spectra are selected. 

Data-centric Deep Nonnegative Matrix Factorization (DC-DNMF)\cite{de_handschutter_consistent_2023} minimizes the revised data-centric D-NMF objective function with a projected gradient optimization. Notably, the objective function used by DC-DNMF is essentially the same as that used by BLUTH, with the exception of the HASC and constraints on the endmember spectra. The layer-wise structure is defined before training, unlike H2NMF and BLUTH, and the number of layers are chosen to follow the architecture found by the BLUTH algorithm on the same scene. The MatLab implementation of DC-DNMF is used.

\subsection{Binary updates with additional endmembers}
\label{ap:binary_updates}
\renewcommand{\theequation}{S\arabic{equation}}

For scenes with more than two endmembers, binary unmixing can be applied by updating each endmember separately in a {\it 1-v-rest} configuration.
Then the abundance of endmember $\zeta$ is:
\begin{equation}
    a_{\zeta} = \frac{(\bm{y}-\bm{s}_{\slashed{\zeta}})^{\intercal}(\bm{s}_{\zeta}-\bm{s}_{\slashed{\zeta}})-\gamma \|\bm{a}_{\slashed{\zeta}}\|_{2}^{2}\|\bm{a}_{\slashed{\zeta}}\|_{1}^{-2}}{\|(\bm{s}_{\zeta}-\bm{s}_{\slashed{\zeta}})\|_{2}^{2} - \gamma(1+\|\bm{a}_{\slashed{\zeta}}\|_{2}^{2}\|\bm{a}_{\slashed{\zeta}}\|_{1}^{-2})}
    \label{eq:quasibinary_abundance_update}
\end{equation}
where $\bf{a}_{\slashed{\zeta}}$ indicates the abundance vector with endmember $\zeta$ removed, and $\bm{s}_{\slashed{\zeta}}$ is the spectrum implied by the relative abundances of the endmembers other than $\zeta$. In the case when $\|\bm{a}_{{\slashed{\zeta}}}\|_{1}=0$, it is assumed that $\bm{s}_{\slashed{\zeta}}$ is the average of the spectra of the endmembers $\neq \zeta$. 

\subsection{Batchable Archetype Analysis and Pure Pixel Analysis}
\label{ap:spectra}

Archetypal analysis (AA) uses the same objective function as NMF, but constrains the matrix $S$ so that it must be a convex sum of the data points themselves \cite{zouaoui_entropic_2023}:
\begin{align}
\min_{S, A}\|Y - SA\|_{\text{F}}^{2}, 
\label{eq:aa_objective.}
\end{align}
with the constraint 
\begin{align}
    S = YX 
    \label{eq:AA_con}
\end{align} where all the entries of $X$ are non-negative and its columns sum to one. 
The constraint implies that each endmember is a weighted average of pixels in the scene. 
One challenge of AA is that $X$ has size $N \times P$, which means that it is slow to update and does not straightforwardly permit batching during the training process. 
Here we introduce a new update rule that permits batching.

Instead of directly updating $X$, $S$ is updated in a way that implicitly satisfies equation \ref{eq:AA_con}.
As in NMF, AA typically minimizes $S$ and $A$ though alternating minimazation. 
Because the update of $A$ can be performed by any variant of NMF, only the update of $S$ is considered here.
Furthermore, we consider cyclic block coordinate descent over the different endmembers, so that each is column of $S$ is updated sequentially. 

Now, consider the spectra of one pixel from the data, $\bm{z}$, and an endmember $\bm{s}_{\zeta}$ that satisfies equation \ref{eq:aa_objective.}.
Then the part of the objective function which depends on $S$ can be re-written so that:
\begin{align}
\label{eq:aa_soln}
\min \|Y - &(S+b U)A\|_{\text{F}}^{2}  \\
=\min&\|E\|_{\text{F}}^{2} - 2b \text{Tr}(E A^{\intercal} U^{\intercal}) + b^{2} \|UA\|_{\text{F}}^{2} \nonumber
\end{align}
where the $\zeta$th column of $U$ is $\bm{u}_{\zeta}=\bm{z}-\bm{s}_{\zeta}$ while the others are zero, $E=Y-SA$, and $b \in (0,1)$.
Then, $(1-b)\bm{s}_{\zeta} + b \bm{z}$ satisfies the AA constraint on endmember spectra as long as $\bm{s_{\zeta}}$ already satisfied the constraint. 
For a given $u_{\zeta}$, the $b$ which minimizes the objective function is then:
\begin{align}
    b = \frac{\text{tr}(E A^{\intercal} U^{\intercal})}{\|UA\|^{2}_{\text{F}}}.
    \label{eq:aa_beta}
\end{align}
Computation can be accelerated by rewriting the equation to neglect the empty columns of $U$. In addition, pixel $\bm{x}$ is dropped from $Y$ and $A$ to avoid biasing the estimate of $b$ for small batch sizes. If $b$ calculated according to equation \ref{eq:aa_beta} falls outside the allowed range, it is set to 0. Once $b$ is found for each pixel in the batch, the pixel which minimizes the objective function is selected.

A related technique, Pure Pixel Analysis (PPA), is defined restricting AA further so that only spectra found in the scene can be selected.  
Spectra which satisfy PPA thus satisfy AA, although the inverse is not true.
PPA imparts even more sparsity than AA because it implies that for each endmember there is at least one pixel in the dataset for which an abundance of 1 would minimize the objective function.

The objective function is then computed for each pixel spectrum in the batch.
The spectrum is initially set to be the one that minimizes the objective function.
If the selected spectrum is identical to a current endmember spectrum, then the spectrum which leads to the next-smallest objective function is selected.
Then, as long as the batch size is larger than the number of endmembers, the new endmember spectrum will not be identical to one of the previous endmember spectra. 

Note that AA performed in this way is guaranteed not to increase the objective function evaluated on the batch used, as $b=0$ for all pixels if none of them can decrease the objective function. 
PPA, on the other hand, can increase the objective function because it always selects a new spectrum, and there may not be a spectrum in the batch which does not increase the objective function. 

\subsection{BLUTH abundance updates}
\label{ap:updates} 

Cyclic block coordinate (gradient) descent is used to train the parameters that a BLUTH uses to partition each pixel into abundances. 
The abundances at any level can be written as a product of the abundances of levels above it (eq. \ref{eq:abundance_product}).
The weights $\bm{w}_{\zeta}^{T}$ and the offset $d_{\zeta}$ can thus be set independently for each node $\zeta$, while the others remain fixed. 
First the direction of the gradient at each node is calculated, and then the step size is determined by a line search. 

The regularized objective function at level can be written as a sum over all pixels:
\begin{align}
\sum_{m,n} \bigg[ \mu_m  \mathbf{u}_{mn} ^{2} - \gamma_m \mathbf{a}_{mn}^{2} \bigg],
\end{align}
where $\mathbf{u}_{m n}=\mathbf{y}_{n}-S_{m}\mathbf{a}_{mn}$ is the the discrepancy between the observed $n$th pixels and its representation at the $m$th level, and $\mathbf{a}_{mn}$ is its abundance ($n$th column of $A_{m}$).

Taking a derivative of the first term with respect to the weights at the $\zeta$ node in the $l$th level (m$\geq$l), we find that
\begin{align}
    \frac{d \mathbf{u}_{mn}^{2}}{d \mathbf{w}_{\zeta}} &=  -2\mathbf{u}_{mn}\bigg(S_{m}\frac{d \mathbf{a}_{mn}}{d \mathbf{w}_{\zeta}}\bigg). 
\end{align}
The abundances change as a function of the weights:
\begin{align}    
     \frac{d \mathbf{a}_{mn}}{d \mathbf{w}_{\zeta}}  &=  \frac{\mathbf{y}_{n}}{2}(\mathbf{a}_{ln}\mathfrak{A}_{lmn})^{\intercal}\Phi'_{\zeta n}, %
\end{align}
where $\mathfrak{A}_{lmn}\in \mathbb{R}^{k \times k}$ is a diagonal matrix defined elementwise as: 
\begin{align}
(\mathfrak{a}_{lmn})_{\eta} = \pm\prod_{\beta\in(\zeta,\eta]} x_{\beta n}.
\end{align}
The term $(\mathfrak{a}_{lmn})_{\eta}$ is positive if the endmember $\eta$ is on the side of the positive descendants of node $\zeta$, and negative if it is on the negative side, and $\Phi'_{\zeta}$ is one if the argument of the cReLU operator is between 0 and 1, and zero otherwise.

The effective spectra of the positive and negative side of node $\zeta$ at level $l$ can then be written as:
\begin{equation}
\bm{s}^{\pm}_{[\zeta\rightarrow m] n} = \sum_{\alpha\in C^{\pm}}\bm{s}_{\alpha n}\bigg(\prod_{\beta\in (\zeta,\alpha]} x_{\beta n}\bigg).
\end{equation}

After simplification, it can be seen the the update depends only on the error, the different between the effective spectra, the abundance, and the pixel spectrum, activated by $\Phi'$:
\begin{align}
    \frac{d \mathbf{u}_{mn}^{2}}{d \mathbf{w}_{\zeta}} &=  -\bigg[\mathbf{u}^{\intercal}_{mn}\big(\mathbf{s}_{[\zeta\rightarrow m] n}^{+}-\mathbf{s}_{[\zeta\rightarrow m] n}^{-}\big)\bigg]a_{\zeta n}\bm{y}_{n}\Phi'_{\zeta n}. 
    \label{eq:consistency_grad}
\end{align}

The gradient that originates from the regularization term is calculated similarly:
\begin{align}
    \frac{d \bm{a}_{mn}^{2}}{d \mathbf{w}_{\zeta}} &=  2\bm{a}^{T}_{mn}\frac{d \bm{a}_{mn}}{d \mathbf{w}_{\zeta}}, \\
    &=\bm{a}^{T}_{mn}(\bm{a}^{+}_{[\zeta\rightarrow m] n} -\bm{a}^{-}_{[\zeta\rightarrow m] n} )a_{\zeta n}\bm{y}_{n}\Phi'_{\zeta n},
    \label{eq:sparsity_grad}
\end{align}
where
\begin{equation}
\bm{a}^{\pm}_{[\zeta\rightarrow m] n} = \sum_{\alpha\in [\zeta\rightarrow m]^{\pm}}a_{\alpha n}\hat{\bm{a}}_{\alpha}\bigg(\prod_{\beta\in (\zeta,\alpha]} x_{\beta n}\bigg),
\end{equation}
where the $\hat{\bm{a}_{\alpha}}$ are used to distinguish the directions that the different endmembers take on in the vector. 

The full gradient with respect to $\bm{w}_{\zeta}$ is then found by summing both terms of the of the objective function, appropriately weighted, over all pixels and all levels below $\zeta$.
The gradient determines the direction of the update step, $\hat{v}_{\zeta}$. 
The magnitude is determined by inserting $b\hat{v}_{\zeta}$ into the objective function and differentiating with respect to $b$.
The above equations can be re-used after accounting for the fact that the derivative of $x_{\zeta}$ with respect to $b$ is a scalar:
\begin{align}
    \frac{dx_{\zeta n}}{db} =  \frac{\mathbf{y}_{n}\cdot \mathbf{\hat{v}}}{2} \Phi_{\zeta n}'.
\end{align}
The derivative with respect to $b$ is then linear except at $2N$ discontinuities. 
Thus, the value of $b$ can be chosen exactly in order to minimize the objective function according to the numerical integration technique described below. 
Therefore, it is not necessary to set a learning rate, reducing the number of hyperparameters. 

\subsection{Integration of piecewise linear derivative}
\label{ap:numintegration}

The objective function is quadratic in $\mathbf{w}_{\zeta}$ except at the points where $(\mathbf{w}_{\zeta}^\intercal\bm{y}_n - d + 1)=0$ for some pixel $n$.
At these points, the function is continuous but the derivative is not uniquely defined.
Note that the gradient can be written as a sum over all the individual pixels (equations \ref{eq:consistency_grad} and \ref{eq:sparsity_grad}).
The factor of $\Phi_{\zeta n}'$ acts as an activation function for whether individual pixels contribute to the gradient. 

Numerical integration is then used to determine where, along the direction of a vector $\hat{v}$ the gradient vanishes. 
First, all the magnitudes $b$ at which there is a zero-intercept of a pixel are calculated. 
Then, using the initial values of $\Phi_{\zeta n}'$, an initial estimate of the $b$ at which the derivative vanishes is computed. 
If no zero-intercepts fall between $b=0$ and the estimated minimum, then $b$ is set to that value for the update step. 
Otherwise, the derivative is integrated up to the nearest zero-intercept and updated by accounting for changes in the activation functions. 
The process stops when the derivative becomes positive, and $b$ is set to the value at which the it vanishes. 
However, if there is a discontinuity in the derivative such that it is never 0, then it is set to a value just below the discontinuity.

\subsection{Unmixing by Annealing}
\label{ap:annealing}

Annealing is a technique for finding a minimum of an objective function by dynamically altering the update process through a decreasing parameter, typically called temperature ($T$, a scalar) in analogy with metallurgical annealing \cite{metropolis_equation_1953,kirkpatrick_optimization_1983}). 
Deterministic annealing NMF (DA-NMF) directly modifies the objective function by adding a penalty term \cite{peng_nonnegative_2012}:
\begin{equation}
G_{\text{DA}}(S,A,T) = G_{0}(S,A) + T G_{s} (A),
\end{equation}
Where $G_{0}$ is the original objective function and $G_{s}$ is a sparsity-reducing penalty term. 
In the limit of $T\rightarrow \infty$, the objective function becomes convex with a global minimum consisting of equal abundances of all endmembers at each pixel. 
The parameter $T$ is then updated following an exponential decay equation determined by the initial conditions and a decay rate. 
In the limit $T\rightarrow0$, the original objective function is recovered, so the minimum recovered by DA-NMF is also a minimum of the original objective function. 
In this way, non-convexity is slowly added to the objective function, so that the final minimum smoothly evolves from the global minimum of the simplified convex objective function, which can be found uniquely. 

The original implementation of DA-NMF performed comparatively well on the Urban scene \cite{peng_nonnegative_2012}, but some challenges were encountered in re-creating it for these tests, particularly with respect to the ASC. 
Since some changes were necessary, the method has been re-developed as Deterministic Annealing Archetypal Analysis (DAAA). 
DAAA proceeds by alternating minimization, as most unmixing algorithms do.
Although the original DA-NMF proposes a sparsity penalty proportional to $-\sum_{p} a_{p} \log a_{p}$, which would correspond to entropy if the abundances were replaced with probabilities, DAAA uses the L2 sparsity penalty because it can be linearized for exact block updates, as in SMUG.  
At each value of $T$, the abundance matrix is updated using equation \ref{eq:quasibinary_abundance_update} and the spectral matrix is updated using equation \ref{eq:aa_beta}.
The algorithm ends after a pre-determined number of iterations. 

Instead of directly altering the objective function, simulated annealing alters how the solution space is explored. 
Because PPA discretizes the process of determining the endmember spectra, it is well suited for simulated annealing. 
In simulated annealing, a random change to the system is proposed and it is accepted or rejected according to how it affects the objective function. 
In Simulated Annealing Pure Pixel Analysis (SAPPA), the $n$th pixel spectrum is proposed as the spectrum for the $\zeta$th endmember with probability:
\begin{equation}
    p_{\zeta n} = \frac{a_{\zeta n}}{\sum_{j=0}^{N} a_{\zeta j}}.
\end{equation}
Once the proposed spectrum is selected, the abundances are updated according to equation \ref{eq:quasibinary_abundance_update} (with $\gamma=0$).
The update step is then accepted with probability:
\begin{equation}
    e^{-(G_f-G_i)/T},
\end{equation}
where $G_i$ is the value of the objective function before the proposed update, and $G_f$ is the updated value. 
Therefore, proposed update steps that decrease the objective function are always accepted, but there is also non-vanishing probability of accepting update steps that increase the objective function. 
This allows simulated annealing to escape local minima and, ideally, find the global minimum. 
\ifCLASSOPTIONcaptionsoff
  \newpage
\fi

\subsection{Step Sizes and Setpoints in Sparsity Modulation}
\label{ap:SMsssetpoints}

Rather than setting one sparsity penalty for the whole optimization process, SMUG dynamically adjusts a sparsity penalty according to the state of the system and the modality that the optimization algorithm is in. The changes to the sparsity penalty are parametrized by the sparsity step size $\delta_{\gamma}$, or how much the penalty changes between steps of the optimization algorithm. 

The SpM begins by determining the initial maximum value of $\gamma$ to be applied as:
\begin{equation}
    \gamma_{\text{max}}=\frac{G_{ll}}{(\sum_{\zeta,n} a^{2}_{\zeta n})/N - P^{-1}},
    \label{eq:gain_step_size}
\end{equation}
where $G_{ll}$ is the data-centric term at the lowest layer from the objective function, and the denominator adjusts for the current sparsity of $A$.
Each step of the SpM gain adjustment then adds $\gamma_{\text{max}}/n_{\text{runs}}$ to $\gamma$.
After each set of $n_{\text{runs}}$, $\gamma_{\text{max}}$ is adjusted depending on how much the pure pixel proportion (PPP) has changed.
As PPP changes more, the gain is adjusted more slowly. Before there is any change to the PPP, it is multiplied $2\times$, but close to the setpoint, it changes only $1.1\times$. 
   
In ShM, the step size is initialized to $\gamma_{\text{max}}$ above, but it is increased linearly instead of multiplicatively. 
After the relaxation step in ShM, the $\gamma$ calculated by equation \ref{eq:gain_step_size} is added again to the setpoint. 

In DeSM the baseline gain is initialized to $-G_{ll}$ in equation \ref{eq:gain_step_size}. For the $i$th step, $\gamma$ is $-G_{ll}/i$. 

\vspace{15pt}

\begin{table}[h]
\centering
\setlength{\tabcolsep}{3pt}
\begin{tabular}{r|r|rr|rr|rr|rr}
& \multicolumn{1}{c}{$\epsilon$}  &
\multicolumn{2}{c|}{Trees} & \multicolumn{2}{c|}{Water} & \multicolumn{2}{c|}{Dirt} & \multicolumn{2}{c}{Road} \\
 &  & PPA & AA & PPA & AA & PPA & AA & PPA & AA \\
 \hline
IoU & 0.00 &  \bf{0.916} & 0.914 & 0.548 & 0.576 & 0.704 & 0.705 & 0.095 & 0.102 \\
& 0.06 &  0.905 & 0.909 & 0.525 & 0.554 & 0.692 & 0.679 & 0.032 & 0.025 \\
$\rightarrow $& 0.11 &  0.792 & 0.819 & 0.913 & 0.912 & 0.756 & 0.768 & 0.676 & 0.659 \\
& 0.23 &  0.821 & 0.819 & 0.934 & 0.931 & 0.788 & 0.770 & 0.731 & 0.689 \\
& 0.45 &  0.848 & 0.843 & 0.933 & \bf{0.935} & \bf{0.800} & 0.788 & \bf{0.740} & 0.707 \\
& 1.00 &  0.791 & 0.767 & 0.739 & 0.704 & 0.641 & 0.737 & 0.569 & 0.525 \\
\hline
SA & 0.00 &  \bf{2.89} & 2.96 & 7.93 & 7.198 & 3.73 & 4.20 & 33.22 & 34.09 \\
& 0.06 &  3.04 & 2.90 & 15.33 & 10.77 & 3.73 & 5.77 & 61.01 & 59.10 \\
$\rightarrow$& 0.11 &  6.23 & 5.75 & 3.43 & \bf{2.75} & \bf{1.52} & 1.77 & 2.58 & 3.47 \\
& 0.23 &  5.04 & 6.39 & 4.47 & 3.34 & \bf{1.52} & 2.04 & 2.33 & 3.48 \\
& 0.45 &  4.85 & 5.81 & 4.47 & 3.35 & 1.85 & 2.83 & \bf{1.31} & 2.10 \\
& 1.00 &  4.31 & 4.65 & 4.44 & 5.33 & 7.20 & 5.03 & 4.40 & 2.53 \\
\end{tabular}
\vspace{1pt}
\caption{Jasper Ridge metrics with sorted by partial normalization exponent. The largest IoU (smallest SA) for each endmember is in a bold font. The arrow shows the level of partial normalization that was selected based on these experiments. The intermediate $\epsilon$ are calculated from $\nu$=1/8,1/4,1/2 and 1 (equation \ref{eq:epsilon})}
\label{tab:Jmetrics}
\end{table}

\vspace{15pt}

\begin{figure}[h]
\centering
\includesvg[width=0.48\textwidth]{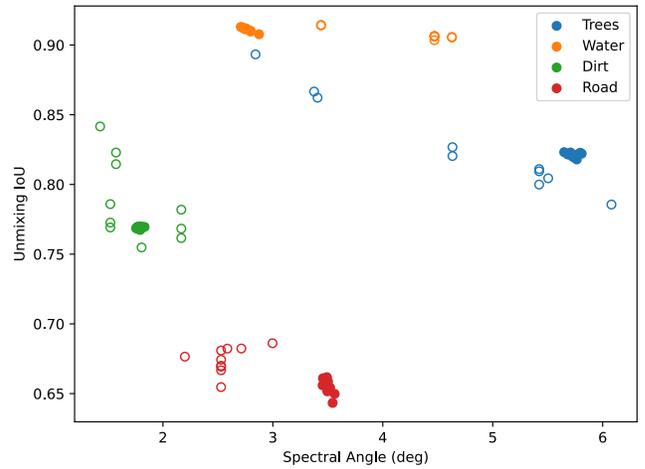}
\caption{The Jasper scene unmixed 10 times with minibatching and a batch size of half the pixels. Open markers correspond to PPA and closed markers correspond to AA in the fine-tuning stage.}
\label{fig:J_unmixing_consistency}
\end{figure}

\begin{table*}
\begin{tabular}{ll|rrrrrrrrrrrr}
Scene & Class & BLU-AA & BA+2 & BLU-PPA & BP+2 & DAAA & SAPPA & EDAA & QMV & MiSiC & MSNet & H2NMF & DC-DMF \\
\hline
Samson & Soil & 0.52 & 0.71 & 0.64 & 0.66 & \bf 0.45 & \it 0.50 & 1.65 & 4.54 & 10.87 & 2.87 & 2.88 & 1.52 \\
 & Tree & 1.89 & 1.85 & \it 1.79 & \bf 1.54 & 2.13 & 1.88 & 1.98 & 7.68 & 3.46 & 1.96 & 4.90 & 2.68 \\
 & Water & 1.79 & 6.77 & \it 1.74 & 6.25 & 2.55 & 4.21 & \bf 1.31 & 17.44 & 9.19 & 1.81 & 6.15 & 5.99 \\
\hline
Jasper & Tree & 5.75 & 5.25 & 6.24 & 4.72 & 4.26 & 5.50 & 4.24 & 13.31 & 10.86 & \it 3.55 & \bf 2.58 & 10.06 \\
 & Water & \it 2.75 & 4.39 & 3.44 & 3.44 & 10.99 & 5.65 & 2.79 & 15.99 & 12.68 & \bf 1.86 & 8.38 & 10.60 \\
 & Dirt & \it 1.78 & 7.37 & \bf 1.52 & 6.57 & 4.63 & 1.93 & 2.78 & 11.75 & 43.41 & 10.99 & 13.57 & 40.15 \\
 & Road & 3.48 & 9.65 & \it 2.59 & 2.73 & 59.88 & \bf 2.29 & 3.01 & 36.69 & 19.28 & 22.84 & 8.21 & 18.85 \\
\hline
Apex & Road & 7.11 & 8.40 & \bf 6.16 & \it 6.42 & 39.41 & 11.89 & 13.55 & 56.22 & 51.10 & 45.19 & 13.25 & 54.11 \\
 & Tree & 7.45 & 9.81 & 8.52 & 9.37 & \bf 2.09 & \it 6.65 & 7.73 & 16.97 & 37.22 & 7.59 & 9.52 & 30.83 \\
 & Roof & 6.91 & \it 3.72 & 6.52 & \bf 3.62 & 8.55 & 7.53 & 7.85 & 8.05 & 7.05 & 5.93 & 5.33 & 10.80 \\
 & Water & \it 2.50 & 2.61 & 3.26 & 2.69 & 2.80 & 3.26 & \bf 2.09 & 9.79 & 3.08 & 2.55 & 2.81 & 3.54 \\
\hline
Urban & Road & \bf 3.04 & 4.33 & \it 3.31 & 3.43 & 6.43 & 5.09 & 5.53 & 9.18 & 32.34 & 8.26 & 8.07 & 12.99 \\
 & Grass & \bf 2.08 & 2.79 & 2.51 & 3.11 & 2.75 & 3.17 & \it 2.11 & 7.26 & 15.82 & 8.22 & 4.34 & 12.73 \\
 & Tree & 2.89 & 4.25 & 3.20 & 3.66 & 4.34 & \it 2.58 & 2.90 & 9.47 & 10.94 & 2.59 & \bf 1.73 & 6.48 \\
 & Roof & 16.75 & \it 3.21 & 15.91 & \bf 2.42 & 18.13 & 13.55 & 11.14 & 18.27 & 27.68 & 7.11 & 6.08 & 32.93 \\
\hline
Urban & Road & \it 4.67 & 6.73 & \bf 2.39 & 5.67 & 10.52 & 6.48 & 5.12 & 19.61 & 49.31 & 29.83 & 6.08 & 23.65 \\
 & Grass & 4.60 & 2.79 & 3.68 & 2.62 & 7.19 & 5.43 & \it 2.09 & 14.99 & 27.68 & 9.20 & \bf 1.53 & 10.39 \\
 & Tree & 8.25 & 9.58 & 8.12 & 9.63 & \bf 5.06 & 8.01 & 8.87 & 22.50 & 16.57 & 7.43 & \it 6.20 & 11.18 \\
 & Roof & 6.14 & 4.55 & 7.77 & \it 3.85 & 18.37 & \bf 3.06 & 13.81 & 20.23 & 39.71 & 8.17 & 4.09 & 32.84 \\
 & Metal & 46.88 & 52.83 & 47.32 & 50.90 & 47.74 & 46.87 & \bf 4.44 & \it 8.00 & 18.25 & 10.30 & 21.33 & 14.80 \\
 & Dirt & 7.06 & \it 3.85 & 6.07 & 3.86 & \bf 2.85 & 5.05 & 13.86 & 24.28 & 36.96 & 13.57 & 9.45 & 32.80 \\
\hline
WDC & Grass & 4.57 & 4.09 & 4.76 & \it 3.74 & 19.24 & \bf 2.92 & 7.40 & 33.34 & 35.71 & 7.64 & 9.52 & 20.89 \\
 & Tree & 15.09 & 10.14 & 15.34 & 6.72 & 9.74 & \it 4.15 & \bf 2.10 & 20.34 & 33.64 & 15.47 & 7.24 & 28.59 \\
 & Road & 17.87 & 5.33 & \it 2.33 & \bf 1.68 & 34.54 & 8.17 & 4.63 & 40.10 & 31.06 & 24.47 & 5.94 & 16.84 \\
 & Roof & 5.48 & \it 3.04 & 27.08 & \bf 2.28 & 15.58 & 51.31 & 50.93 & 7.49 & 30.42 & 24.12 & 27.58 & 47.05 \\
 & Water & \it 1.32 & 1.33 & 1.93 & 2.01 & 4.40 & 2.24 & \bf 1.17 & 12.90 & 21.79 & 1.68 & 1.97 & 3.58 \\
 & Trail & 5.81 & 6.49 & 4.41 & \it 3.53 & 5.57 & 3.67 & \bf 2.11 & 6.45 & 28.00 & 8.58 & 5.25 & 26.92 \\
\hline
SMS & Moss & 3.89 & 1.84 & \bf 1.41 & \it 1.56 & 2.77 & 2.81 & 8.85 & 25.96 & 11.53 & 8.59 & 2.61 & 5.13 \\
 & Pebbles & \bf 0.32 & 1.26 & 0.97 & 1.69 & \it 0.70 & 0.83 & 5.54 & 14.49 & 2.96 & 9.41 & 18.53 & 8.95 \\
 & Sticks & 12.29 & \bf 4.12 & 16.39 & \it 5.40 & 24.67 & 26.37 & 17.00 & 17.17 & 11.77 & 19.08 & 21.94 & 25.15 \\
 & Leaves & 2.61 & 1.79 & 2.18 & 2.41 & \it 0.80 & \bf 0.27 & 1.64 & 2.19 & 6.31 & 2.76 & 4.98 & 2.25 \\
\hline
CMS & Moss & \it 0.88 & 1.09 & 1.46 & 1.37 & 1.31 & 1.45 & 10.66 & 1.96 & 7.16 & 10.17 & \bf 0.58 & 4.40 \\
 & Pebbles & 6.02 & 4.03 & 6.37 & \it 3.90 & 7.80 & 4.64 & 28.84 & 12.72 & 15.31 & 8.38 & \bf 0.69 & 28.49 \\
 & Sticks & 13.66 & \it 5.88 & 14.06 & \bf 5.14 & 12.14 & 13.79 & 18.31 & 25.82 & 39.90 & 18.05 & 22.13 & 25.71 \\
 & True V & 1.79 & 1.75 & \it 1.56 & 2.07 & \bf 1.07 & 3.08 & 4.24 & 3.04 & 13.77 & 12.14 & 2.46 & 10.22 \\
 & False V & 4.07 & 1.69 & 4.25 & 2.35 & 3.47 & 4.25 & 2.99 & \it 1.19 & 20.32 & 1.30 & \bf 0.74 & 11.13 \\
\hline
\end{tabular}
\caption{The spectral angles of the endmembers calculated in each scene relative to the manually-labelled endmembers in degrees. The smallest spectral angle for each endmember is shown in {\bf bold}, while the second smallest is shown in {\it Italic}.}
\label{tab:SA_test}
\end{table*}
\begin{table*}[h]
\begin{tabular}{ll|rrrrrrrrrrrr}
Scene & Class & BLU-AA & BA+2 & BLU-PPA & BP+2 & DAAA & SAPPA & EDAA & QMV & MiSiC & MSNet & H2NMF & DC-DMF \\
\hline
Samson & Soil & 0.876 & 0.809 & 0.865 & 0.816 & \bf 0.917 & 0.908 & \it 0.912 & 0.744 & 0.520 & 0.833 & 0.846 & 0.865 \\
 & Tree & 0.891 & 0.915 & 0.883 & 0.919 & \bf 0.952 & \it 0.942 & 0.934 & 0.754 & 0.755 & 0.904 & 0.870 & 0.885 \\
 & Water & \bf 0.947 & 0.814 & 0.943 & 0.819 & 0.911 & 0.906 & \it 0.945 & 0.638 & 0.667 & 0.840 & 0.885 & 0.814 \\
\hline
Jasper & Tree & 0.819 & 0.705 & 0.792 & 0.715 & \bf 0.897 & 0.851 & 0.864 & 0.721 & 0.755 & \it 0.881 & 0.589 & 0.679 \\
 & Water & \it 0.912 & 0.826 & \bf 0.913 & 0.873 & 0.551 & 0.893 & 0.910 & 0.555 & 0.688 & 0.885 & 0.899 & 0.669 \\
 & Dirt & 0.768 & 0.422 & 0.756 & 0.459 & 0.699 & \it 0.813 & \bf 0.816 & 0.564 & 0.295 & 0.573 & 0.418 & 0.347 \\
 & Road & 0.659 & 0.397 & \bf 0.676 & 0.602 & 0.044 & \it 0.674 & 0.651 & 0.159 & 0.287 & 0.249 & 0.311 & 0.296 \\
\hline
Apex & Road & 0.446 & \it 0.464 & 0.456 & 0.458 & 0.050 & 0.457 & \bf 0.503 & 0.088 & 0.136 & 0.081 & 0.401 & 0.151 \\
 & Tree & \it 0.825 & 0.763 & 0.819 & 0.773 & 0.678 & 0.822 & \bf 0.830 & 0.600 & 0.589 & 0.611 & 0.786 & 0.729 \\
 & Roof & \it 0.745 & 0.595 & 0.719 & 0.600 & 0.702 & 0.744 & \bf 0.746 & 0.647 & 0.653 & 0.691 & 0.649 & 0.676 \\
 & Water & 0.688 & 0.685 & 0.688 & \it 0.688 & 0.671 & \bf 0.706 & 0.667 & 0.545 & 0.594 & 0.616 & 0.659 & 0.593 \\
\hline
Urban & Road & 0.729 & 0.689 & 0.723 & 0.566 & \it 0.796 & 0.795 & \bf 0.833 & 0.725 & 0.511 & 0.706 & 0.740 & 0.660 \\
 & Grass & 0.704 & 0.637 & 0.702 & 0.616 & 0.748 & \it 0.763 & \bf 0.823 & 0.727 & 0.611 & 0.657 & 0.679 & 0.602 \\
 & Tree & 0.728 & 0.709 & 0.740 & 0.674 & 0.734 & 0.762 & \bf 0.866 & 0.659 & 0.594 & \it 0.795 & 0.790 & 0.691 \\
 & Roof & 0.516 & 0.544 & 0.524 & 0.550 & 0.630 & 0.671 & \it 0.709 & 0.569 & 0.415 & 0.515 & \bf 0.727 & 0.498 \\
\hline
Urban & Road & 0.584 & 0.479 & \bf 0.664 & 0.517 & 0.487 & 0.528 & \it 0.662 & 0.416 & 0.130 & 0.276 & 0.596 & 0.303 \\
 & Grass & 0.731 & 0.667 & 0.720 & 0.662 & \bf 0.783 & \it 0.778 & 0.629 & 0.536 & 0.383 & 0.569 & 0.475 & 0.415 \\
 & Tree & 0.712 & 0.652 & 0.713 & 0.630 & 0.527 & \it 0.769 & 0.759 & 0.514 & 0.538 & 0.723 & \bf 0.874 & 0.645 \\
 & Roof & 0.488 & 0.450 & 0.478 & 0.451 & 0.581 & 0.468 & \it 0.678 & 0.326 & 0.334 & 0.425 & \bf 0.735 & 0.420 \\
 & Metal & 0.007 & 0.018 & 0.008 & 0.023 & 0.052 & 0.030 & \bf 0.227 & \it 0.173 & 0.171 & 0.164 & 0.027 & 0.131 \\
 & Dirt & \it 0.611 & 0.490 & \bf 0.654 & 0.496 & 0.388 & 0.575 & 0.453 & 0.232 & 0.412 & 0.480 & 0.367 & 0.304 \\
\hline
WDC & Grass & 0.472 & 0.458 & 0.469 & 0.472 & 0.503 & 0.500 & 0.457 & 0.384 & 0.291 & 0.443 & \it 0.536 & \bf 0.554 \\
 & Tree & 0.428 & 0.568 & 0.425 & \it 0.671 & 0.601 & \bf 0.692 & 0.671 & 0.458 & 0.417 & 0.398 & 0.042 & 0.123 \\
 & Road & 0.141 & 0.414 & \bf 0.625 & 0.520 & 0.080 & \it 0.617 & 0.551 & 0.079 & 0.154 & 0.113 & 0.365 & 0.194 \\
 & Roof & 0.260 & \it 0.275 & 0.036 & \bf 0.275 & 0.095 & 0.007 & 0.011 & 0.119 & 0.076 & 0.126 & 0.022 & 0.143 \\
 & Water & 0.754 & 0.749 & \bf 0.824 & 0.696 & 0.456 & \it 0.803 & 0.787 & 0.468 & 0.370 & 0.716 & 0.766 & 0.629 \\
 & Trail & 0.586 & 0.585 & 0.584 & \it 0.622 & 0.452 & \bf 0.661 & 0.534 & 0.392 & 0.310 & 0.517 & 0.474 & 0.399 \\
\hline
SMS & Moss & 0.343 & \bf 0.724 & 0.585 & \it 0.699 & 0.362 & 0.494 & 0.297 & 0.145 & 0.337 & 0.263 & 0.337 & 0.421 \\
 & Pebbles & 0.145 & \bf 0.486 & 0.124 & \it 0.219 & 0.146 & 0.151 & 0.101 & 0.097 & 0.102 & 0.068 & 0.141 & 0.102 \\
 & Sticks & 0.014 & \it 0.048 & 0.001 & \bf 0.076 & 0.001 & 0.003 & 0.008 & 0.020 & 0.018 & 0.007 & 0.002 & 0.002 \\
 & Leaves & 0.171 & \bf 0.236 & \it 0.224 & 0.183 & 0.114 & 0.131 & 0.079 & 0.065 & 0.093 & 0.102 & 0.062 & 0.101 \\
\hline
CMS & Moss & 0.668 & \it 0.827 & 0.665 & \bf 0.838 & 0.411 & 0.415 & 0.230 & 0.349 & 0.353 & 0.246 & 0.555 & 0.531 \\
 & Pebbles & 0.148 & \it 0.224 & 0.166 & \bf 0.239 & 0.070 & 0.072 & 0.004 & 0.043 & 0.031 & 0.035 & 0.102 & 0.046 \\
 & Sticks & 0.008 & \it 0.025 & 0.007 & \bf 0.027 & 0.006 & 0.007 & 0.007 & 0.002 & 0.007 & 0.010 & 0.002 & 0.007 \\
 & True V & \it 0.448 & 0.389 & \bf 0.451 & 0.386 & 0.206 & 0.238 & 0.172 & 0.224 & 0.110 & 0.099 & 0.334 & 0.230 \\
 & False V & 0.548 & \bf 0.790 & 0.631 & \it 0.779 & 0.536 & 0.594 & 0.596 & 0.604 & 0.227 & 0.494 & 0.716 & 0.359 \\
\hline
\end{tabular}
\caption{The IoUs of the endmembers calculated in each scene relative to the manually-labelled endmembers. The largest IoU for each endmember is shown in {\bf bold}, while the second largest is shown in {\it Italic}.}
\label{tab:IoU_tests}
\end{table*}

\begin{figure*}
\centering
\includesvg[width=1.00\textwidth]{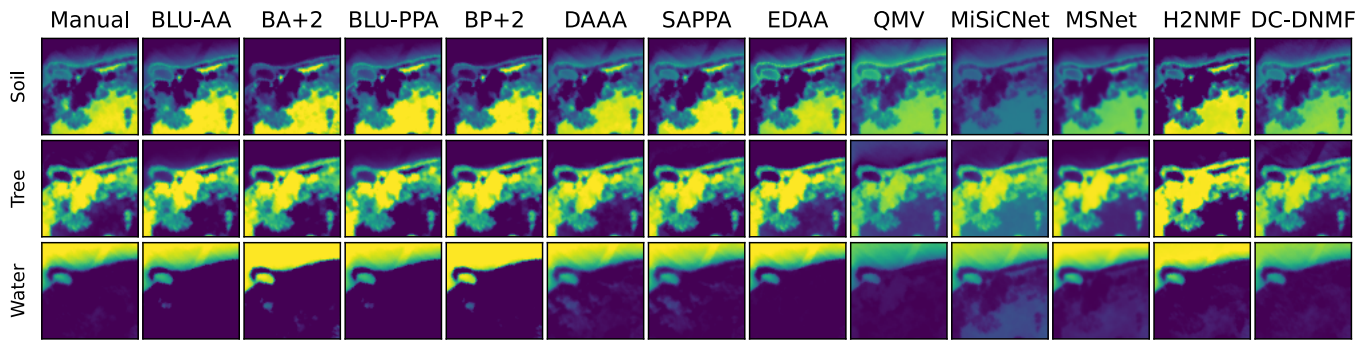}
\caption{The Samson scene unmixed into three endmembers by several algorithms.}
\label{fig:Samson_spatial}
\end{figure*}

\begin{figure*}
\centering
\includesvg[width=1.00\textwidth]{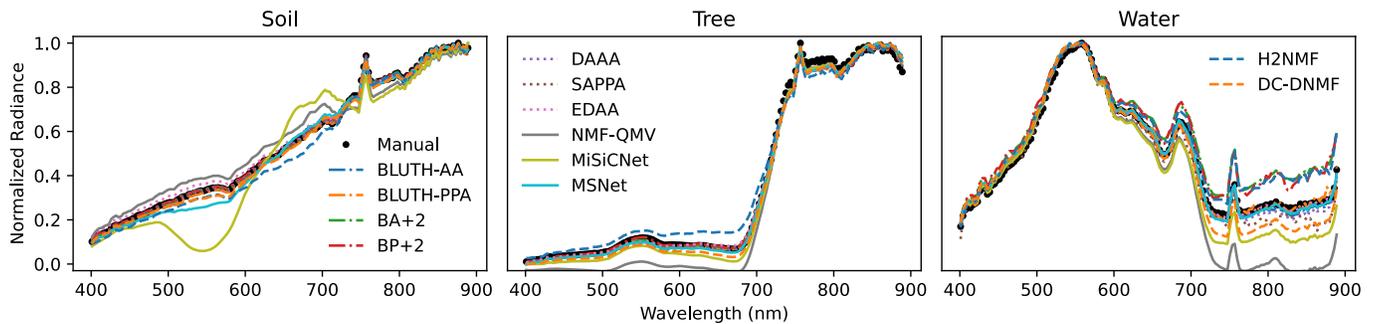}
\caption{The endmember spectra predicted by the unmixing of the Samson scene. Dotted lines correspond to techniques based on AA/PPA and dashed lines correspond to hierarchical techniques. }
\label{fig:Samson_spectral}
\end{figure*}

\begin{figure*}
\centering
\includesvg[width=1.00\textwidth]{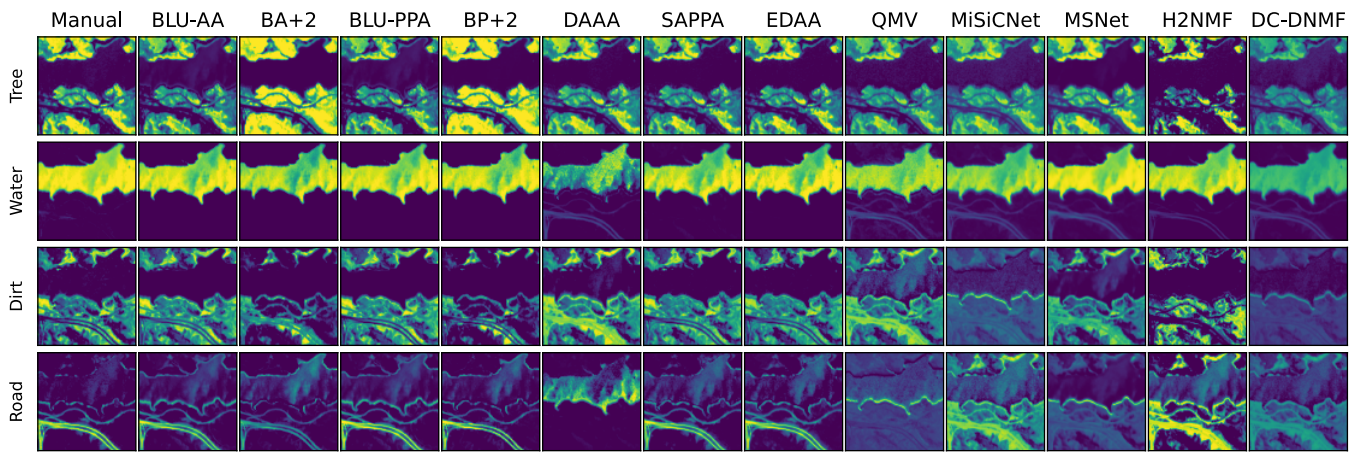}
\caption{The Jasper scene unmixed into 4 endmembers.}
\label{fig:Jasper_spatial}
\end{figure*}

\begin{figure*}
\centering
\includesvg[width=1.00\textwidth]{Figures/Urban4SpatialComparison.svg}
\caption{The Urban scene unmixed into 4 endmembers.}
\label{fig:U4_spatial}
\end{figure*}

\begin{figure*}
\centering
\includesvg[width=1.00\textwidth]{Figures/Urban6SpatialComparison.svg}
\caption{The Urban scene unmixed into 6 endmembers.}
\label{fig:U6_spatial}
\end{figure*}

\begin{figure*}
\centering
\includesvg[width=1.00\textwidth]{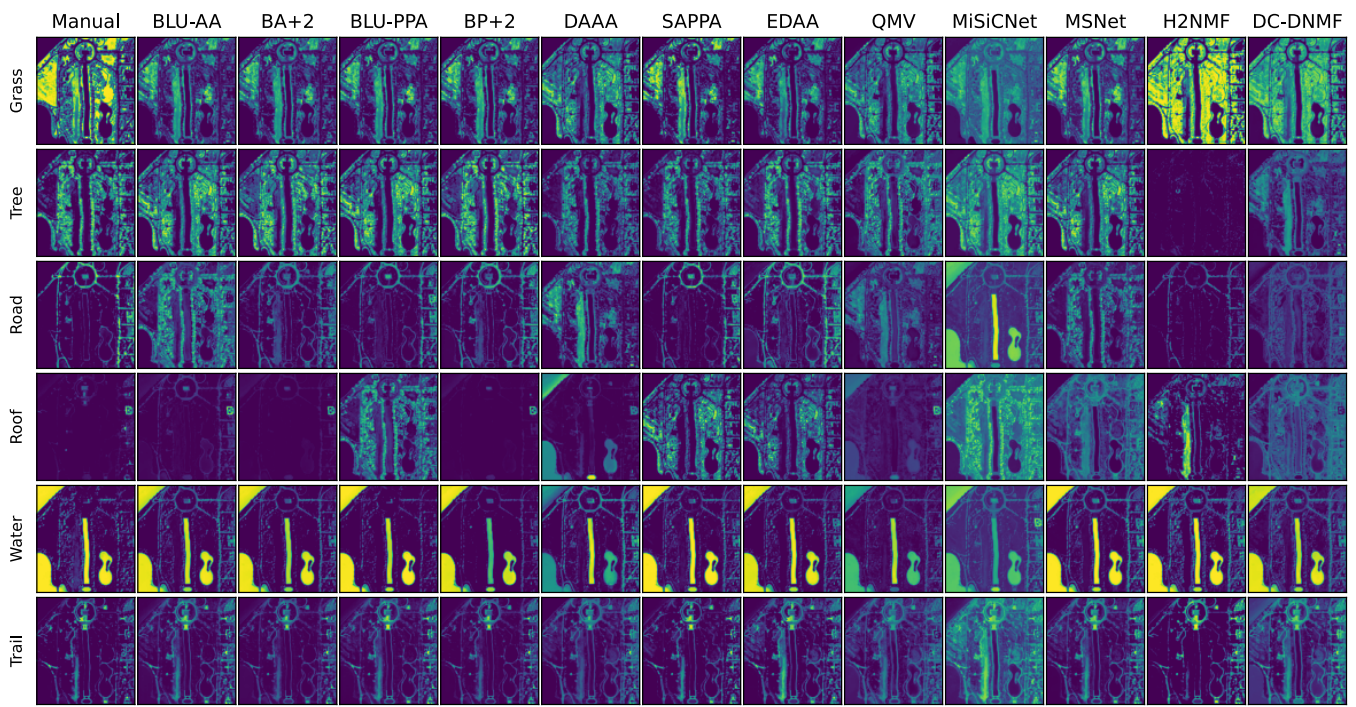}
\caption{The Washington, DC scene unmixed into 6 endmembers.}
\label{fig:WDC_spatial}
\end{figure*}
    \clearpage

\begin{figure*}
\centering
\includesvg[width=1.00\textwidth]{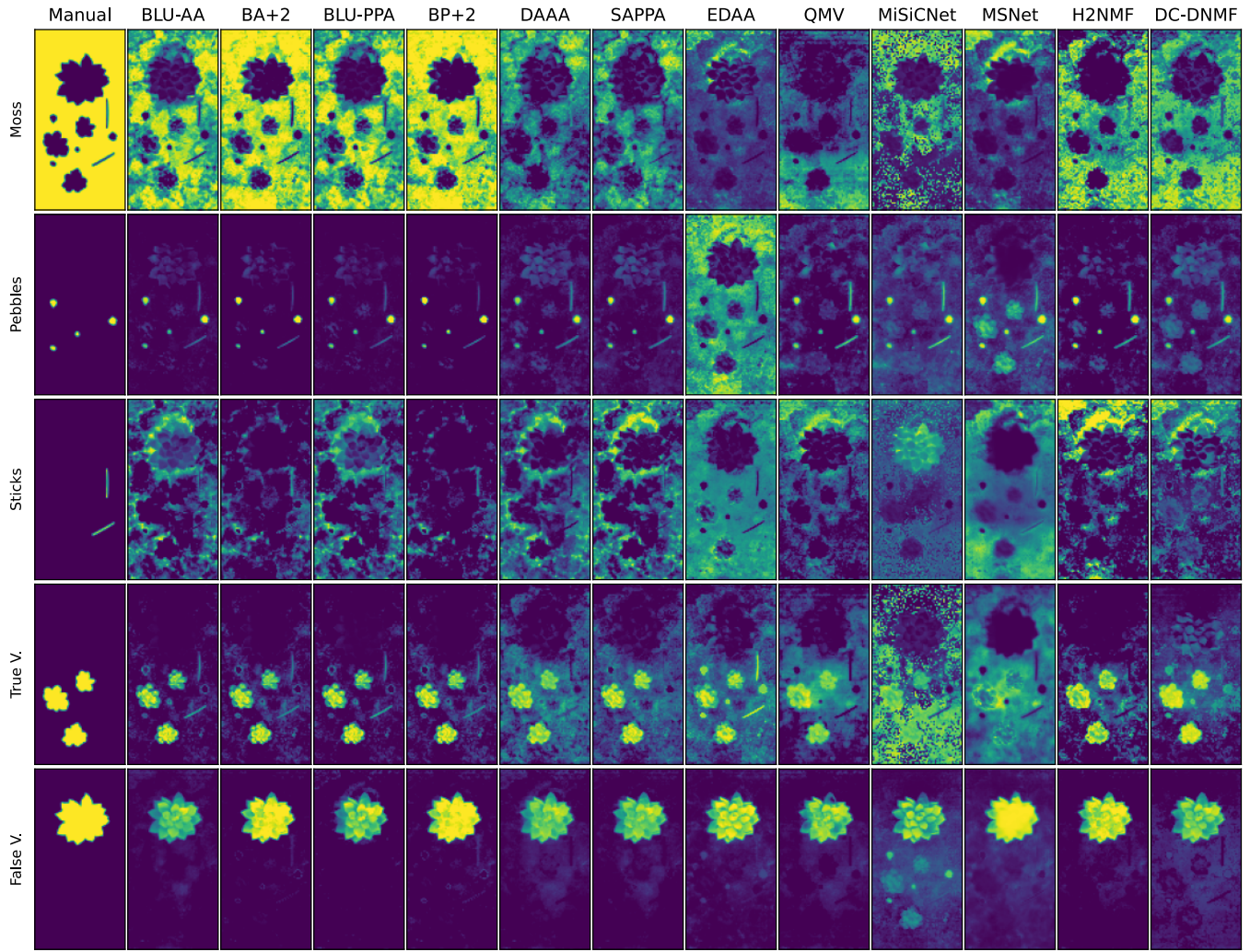}
\caption{The abundances of the RMMS-Complex Mixing Scene unmixed by 12 different techniques.}
\label{fig:CMS_spatial}
\end{figure*}

\end{appendices}

\end{document}